\documentclass[sigconf]{acmart}

\usepackage[utf8]{inputenc} 
\usepackage[T1]{fontenc}    
\usepackage{hyperref}       
\usepackage{url}            
\usepackage{booktabs}       
\usepackage{amsfonts}       
\usepackage{nicefrac}       
\usepackage{microtype}      
\usepackage{xcolor}         
\usepackage{subfigure}
\usepackage{graphicx}
\usepackage{stfloats}

\usepackage{subfigure}
\usepackage{color}
\usepackage{xcolor}
\usepackage{comment}
\usepackage{algorithm,algpseudocode}
\usepackage{rotating}
\usepackage{multirow}
\usepackage{tabularray}

\usepackage{enumitem}
\usepackage{url}
\usepackage{xspace}
\usepackage{subfigure}
\usepackage{amsmath}
\usepackage{tabularx}  
\usepackage{graphicx}
\usepackage{array}
\usepackage{balance}
\usepackage{newtxmath}
\usepackage{makecell}
\usepackage{pifont}

\usepackage{listings}
\usepackage{xcolor}

\definecolor{eclipseStrings}{RGB}{42,0.0,255}
\definecolor{eclipseKeywords}{RGB}{127,0,85}
\definecolor{bittersweet}{rgb}{1.0, 0.44, 0.37}
\colorlet{numb}{magenta!60!black}

\lstdefinelanguage{json}{
    basicstyle=\normalfont\ttfamily,
    commentstyle=\color{eclipseStrings}, 
    stringstyle=\color{eclipseKeywords}, 
    numbers=left,
    numberstyle=\scriptsize,
    stepnumber=1,
    numbersep=6pt,
    showstringspaces=false,
    breaklines=true,
    frame=lines,
    string=[s]{"}{"},
    comment=[l]{:\ "},
    morecomment=[l]{:"},
    literate=
        *{0}{{{\color{numb}0}}}{1}
         {1}{{{\color{numb}1}}}{1}
         {2}{{{\color{numb}2}}}{1}
         {3}{{{\color{numb}3}}}{1}
         {4}{{{\color{numb}4}}}{1}
         {5}{{{\color{numb}5}}}{1}
         {6}{{{\color{numb}6}}}{1}
         {7}{{{\color{numb}7}}}{1}
         {8}{{{\color{numb}8}}}{1}
         {9}{{{\color{numb}9}}}{1}
}

\lstdefinelanguage{yaml}{
  keywords={true,false,null,y,n},
  sensitive=false,
  breaklines=true,
  frame=lines,
  comment=[l]{\#},
  morecomment=[s]{/*}{*/},
  morestring=[b]',
  morestring=[b]",
  showstringspaces=false,
  commentstyle=\color{gray},
  keywordstyle=\color{orange},
  basicstyle=\ttfamily\small,
  stringstyle=\color{blue}, 
  moredelim=[s][\color{blue}]{"}{"}, 
  moredelim=[is][\color{black}]{'}{'}, 
}

\newcommand{\eg}{\emph{e.g.},\xspace}
\newcommand{\ie}{\emph{i.e.},\xspace}
\newcommand{\etal}{\emph{et al.},\xspace}

\setcopyright{none}
\renewcommand\footnotetextcopyrightpermission[1]{} 
\title{Large Action Models: From Inception to Implementation}

\author{Lu Wang}
\authornote{These authors contributed equally to this work. For inquiries, please contact: \{wlu, fangkaiyang, chaoyun.zhang\}@microsoft.com.}
\affiliation{%
  \institution{Microsoft}
  \country{}
}

\author{Fangkai Yang}
\authornotemark[1]
\affiliation{%
  \institution{Microsoft}
  \country{}
}

\author{Chaoyun Zhang}
\authornotemark[1]
\affiliation{%
  \institution{Microsoft}
  \country{}
}

\author{Junting Lu}
\authornote{Work done during internship at Microsoft.}
\affiliation{%
  \institution{Peking University}
  \country{}
}

\author{Jiaxu Qian}
\authornotemark[2]
\affiliation{%
  \institution{Peking University}
  \country{}
}

\author{Shilin He}
\affiliation{%
  \institution{Microsoft}
  \country{}
}

\author{Pu Zhao}
\affiliation{%
  \institution{Microsoft}
  \country{}
}

\author{Bo Qiao}
\affiliation{%
  \institution{Microsoft}
  \country{}
}

\author{Ray Huang}
\affiliation{%
  \institution{Microsoft}
  \country{}
}

\author{Si Qin}
\affiliation{%
  \institution{Microsoft}
  \country{}
}

\author{Qisheng Su}
\authornotemark[2]
\affiliation{%
  \institution{Peking University}
  \country{}
}

\author{Jiayi Ye}
\authornotemark[2]
\affiliation{%
  \institution{Zhejiang University}
  \country{}
}

\author{Yudi Zhang}
\authornotemark[2]
\affiliation{%
  \institution{Eindhoven University of Technology}
  \country{}
}

\author{Jian-Guang Lou}
\affiliation{%
  \institution{Microsoft}
  \country{}
}

\author{Qingwei Lin}
\affiliation{%
  \institution{Microsoft}
  \country{}
}

\author{Saravan Rajmohan}
\affiliation{%
  \institution{Microsoft}
  \country{}
}

\author{Dongmei Zhang}
\affiliation{%
  \institution{Microsoft}
  \country{}
}

\author{Qi Zhang}
\affiliation{%
  \institution{Microsoft}
  \country{}
}

\begin{document}

\begin{abstract} As AI continues to advance, there is a growing demand for systems that go beyond language-based assistance and move toward intelligent agents capable of performing real-world actions. This evolution requires the transition from traditional Large Language Models (LLMs), which excel at generating textual responses, to Large Action Models (LAMs), designed for action generation and execution within dynamic environments. Enabled by agent systems, LAMs hold the potential to transform AI from passive language understanding to active task completion, marking a significant milestone in the progression toward artificial general intelligence.

In this paper, we present a comprehensive framework for developing LAMs, offering a systematic approach to their creation, from inception to deployment. We begin with an overview of LAMs, highlighting their unique characteristics and delineating their differences from LLMs. Using a Windows OS-based agent as a case study, we provide a detailed, step-by-step guide on the key stages of LAM development, including data collection, model training, environment integration, grounding, and evaluation. This generalizable workflow can serve as a blueprint for creating functional LAMs in various application domains. We conclude by identifying the current limitations of LAMs and discussing directions for future research and industrial deployment, emphasizing the challenges and opportunities that lie ahead in realizing the full potential of LAMs in real-world applications.

The code for the data collection process utilized in this paper is publicly available at: {\color{bittersweet}\url{https://github.com/microsoft/UFO/tree/main/dataflow}}, and comprehensive documentation can be found at 
\\{\color{bittersweet}\url{https://microsoft.github.io/UFO/dataflow/overview/}}.

\end{abstract}

\maketitle

\setcopyright{none}
\settopmatter{printacmref=false} 
\renewcommand\footnotetextcopyrightpermission[1]{} 
\pagestyle{plain} 

\section{Introduction} 
In recent years, large language models (LLMs) have demonstrated remarkable advancements across a range of natural language processing (NLP) tasks \cite{wei2021finetuned, brown2020language, yang2023dawn}. These models, often incorporating multiple modalities such as language, vision, and speech, have become foundational in numerous AI-driven applications \cite{thirunavukarasu2023large, rubenstein2023audiopalm, wang2024visionllm, jiang2024xpert}. Their success is evident in systems like question answering in conversational agents \cite{ma2023understanding}, code generation in GitHub Copilot \cite{yeticstiren2023evaluating}, and improved search capabilities in platforms like Bing \cite{thomas2024large}. The key strengths of LLMs—namely their vast knowledge, ability to support multimodal inputs, and capacity for human-like responses—have propelled them to the forefront of AI research \cite{minaee2024large}. Their capability to generalize via zero-shot learning has further expanded the horizons of what AI systems can achieve, making significant contributions to the productivity of both everyday tasks and specialized professional activities. These innovations mark an important milestone on the path toward artificial general intelligence (AGI) \cite{fengfar}.

However, while LLMs excel in generating intricate textual responses, they are often constrained by their inability to directly interact with or manipulate the physical world \cite{wang2024survey}. In many real-world applications, intelligent systems need to perform tasks that go beyond conversational exchanges—tasks that involve tangible actions \cite{gao2024physically}. The maxim ``actions speak louder than words'' \cite{pennycook1985actions} underscores the limitations of purely text-based interactions, as users increasingly expect intelligent agents to go beyond passive responses and engage in real-world actions. For instance, a truly transformative AI assistant could automate tasks in software applications, manage household chores, or even engage with children in meaningful ways. The realization of such capabilities would mark a revolutionary shift in how we integrate AI into our daily lives, enabling widespread automation and augmenting human capabilities across diverse environments \cite{ruan2023tptu}.

Achieving this vision requires LLMs to extend their expertise from language processing to action generation. However, this transition is not straightforward. While leading LLMs from industry giants have demonstrated impressive performance in language-based tasks, they encounter substantial limitations when tasked with action generation \cite{yao2020keep}. Completing a task in the real world involves a sequence of complex steps: accurately understanding user intent, devising a plan, and executing the necessary actions \cite{kalakonda2023action}. Current LLMs may excel at understanding and planning in textual form but often fall short when required to produce actionable outputs. This is particularly true in scenarios that demand precise task decomposition, long-term planning \cite{ding2023everything,zhang2024ruag}, and the coordination of multi-step actions \cite{valmeekam2022large}. Furthermore, LLMs are generally optimized for broad, general-purpose tasks rather than tailored for specific scenarios or environments. This lack of specialization can result in suboptimal performance, especially when interacting with unfamiliar or dynamic environments where adaptive and robust action sequences are essential \cite{ling2023domain}.

These limitations highlight a significant gap in the ability of LLMs to transition from passive understanding to active, real-world engagement. To address these challenges, the development of Large Action Models (LAMs) represents a transformative shift in AI capabilities \cite{he2024words}. Unlike traditional LLMs that primarily focus on text generation and response, LAMs are designed to perform actions in both physical and digital environments. These models are capable of interpreting user intentions from diverse data inputs, automating complex processes, planning for task completion, and interacting with the world via agents. This evolution marks a critical step toward a future where intelligent systems not only comprehend human language but can also translate that understanding into tangible, meaningful actions \cite{zhang2024xlam}.

\begin{figure*}[t]
    \centering
    \includegraphics[width=\textwidth]{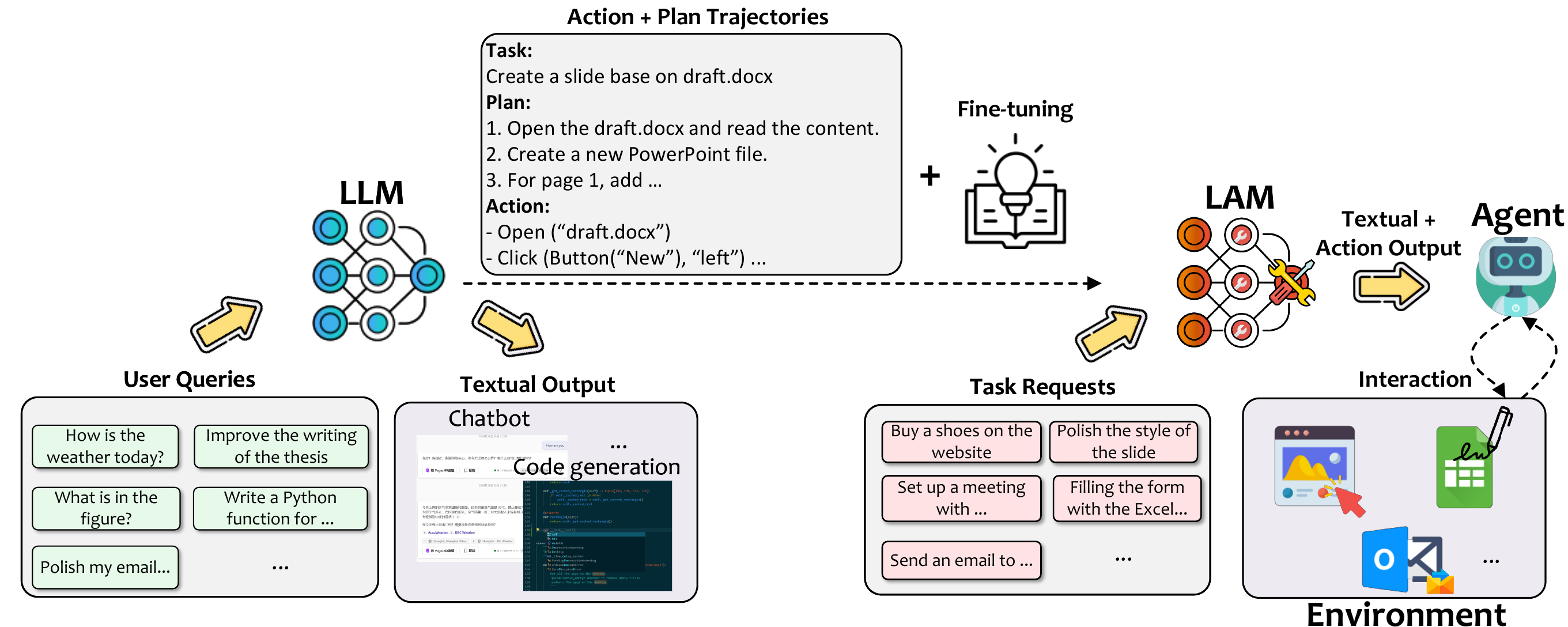}
    \vspace{-2em}
    \caption{The transition from LLMs to LAMs.}
    \label{fig:llm_lam}
\end{figure*}

LAMs are often built upon the foundation of LLMs, but the transition from LLMs to LAMs is neither straightforward nor seamless, as shown in Figure~\ref{fig:llm_lam}. The process of transforming an LLM into a functional LAM involves multiple intricate stages, each requiring substantial effort and expertise. First, it is essential to collect comprehensive datasets that capture user requests, environmental states, and corresponding actions \cite{deng2024mind2web}. These data serve as the basis for training or fine-tuning LLMs to perform actions rather than merely generate text. This stage involves the integration of advanced training techniques that enable the model to understand and execute actions within specific environments \cite{hong2024cogagent}. Once the LAM has been trained, it must be incorporated into an agent system that can effectively interact with its environment. This system typically includes components for gathering observations, utilizing tools, maintaining memory, and implementing feedback loops. These components are critical for ensuring that the LAM can not only execute actions but also adapt its behavior based on real-time feedback and evolving situations \cite{ufo}. The integration of these elements enhances the LAM's capacity to perform tasks autonomously, interact meaningfully with its surroundings, and make decisions that are grounded in the context of its environment.

A final but crucial step in the development of LAMs is evaluation \cite{xie2024osworld}. Before deploying a LAM for real-world applications, it is imperative to rigorously assess its reliability, robustness, and safety. Unlike LLMs, which may be limited to generating text-based outputs, LAMs have the capacity to directly affect their environment through actions. This introduces new risks, as incorrect or inappropriate actions could have significant consequences. Therefore, thorough evaluation processes are essential to ensure that both the LAM and its accompanying agent are capable of making reliable decisions while minimizing potential risks. These evaluations often involve testing the model in a variety of scenarios to ensure that it can generalize across different environments and tasks, as well as effectively handle unexpected situations.

Given the complexity involved in developing LAMs, the purpose of this paper is to provide a comprehensive understanding of LAMs and guide practitioners in transforming an LLM into a functional LAM for real-world applications. To this end, we first present an overview of LAMs, clarifying their distinctions from traditional LLMs and discussing their unique characteristics. By offering this foundational knowledge, we aim to give readers a clear conceptual understanding of LAMs, enabling them to grasp the broader implications of their development and use.

Next, we delve into the practical process of obtaining a LAM from scratch. Using a Graphical User Interface (GUI) agent on Windows OS as an example, we provide a detailed, step-by-step exploration of the entire pipeline—beginning with data collection and preparation, followed by model training, integration, and grounding. This includes how to prepare datasets that capture user requests, environmental states, and actions, as well as how to fine-tune LLMs to generate executable actions rather than text responses. We also demonstrate how to integrate a trained LAM into an agent system, equipping it with tools, memory, and feedback mechanisms to enable dynamic interaction with its environment. The final stages focus on rigorous evaluation, ensuring that the LAM is robust, safe, and capable of handling real-world tasks. While this paper uses the Windows OS as a case study, the methodology outlined can be adapted to other environments, providing a generalizable workflow for obtaining functional LAMs. Finally, we address several limitations and challenges faced by LAMs in both research and industry. While LAMs represent a significant advancement over traditional LLMs, they are still in an early stage of development and present substantial areas for improvement. Issues such as privacy concerns, latency, safety risks, scalability, and ethical considerations all pose challenges that must be addressed for LAMs to be fully realized as practical tools.

The emergence of LAMs represents not merely an incremental advancement over LLMs, but a fundamental shift from passive language processing to active, real-world engagement. By executing actions, LAMs can interact dynamically with both digital and physical environments, marking a transformative milestone in the broader pursuit of AGI. We envision this paper as a foundational guide to LAMs, offering both theoretical insights and practical, actionable steps for creating and deploying LAMs in real-world scenarios.

\section{Large Action Models 101}
Large Action Models (LAMs) represent a significant advancement in artificial intelligence, extending the capabilities of Large Language Models (LLMs)~\cite{zhang2024xlam}. While LLMs are proficient at generating human-like text based on user inputs, LAMs go beyond text generation by performing actions in both physical and digital environments~\cite{zeng2023large}. These models interpret user intentions from various data forms, automate entire processes as per user requirements, plan for task completion, and interact with the world. This evolution signifies a shift from mere language interaction to action sequences that are grounded in real-world contexts.

\subsection{Large Language Models}
LLMs are neural networks with billions to hundreds of billions of parameters, trained on extensive text corpora to address general-purpose language tasks~\cite{kasneci2023chatgpt, xu2022systematic, zhang2024allhands, zhu2023large, liu2024large}. These models demonstrate exceptional capabilities in natural language understanding and generation, allowing them to perform complex tasks such as answering questions~\cite{jiang2021can}, generating code~\cite{zhang2023planning}, and providing human-like textual responses~\cite{dasgupta2022language} with minimal task-specific training, known as zero-shot~\cite{wei2021finetuned} or few-shot~\cite{brown2020language} learning. Unlike traditional language models, which required extensive task-specific data and training, LLMs leverage their vast knowledge base to generalize across diverse tasks with minimal supervision.

While LLMs possess significant language understanding and generation capabilities, they are primarily limited to generating text-based outputs. They excel at interacting with users and generating text, but they lack the ability to directly interface with environments to execute actions. This limitation restricts their applicability in scenarios that require tangible interaction with digital or physical environments.

To extend their utility, LLMs are often embedded within agent frameworks~\cite{wang2024survey}. These agent systems augment LLMs, enabling them to interact with dynamic environments by collecting data from various sources~\cite{xi2023rise}, structuring it into meaningful inputs~\cite{kim2024language}, and prompting the LLM for inference~\cite{xi2023rise}. The agent then interprets the model's output---whether in the form of code~\cite{wang2024opendevin} or tool-based actions~\cite{ruan2023tptu}---and grounds it within the environment by executing actions and collecting feedback~\cite{shinn2024reflexion}. Agents equipped with LLMs typically function in a loop, continuously gathering environmental information, using LLM inference to form plans, executing those plans, and refining future actions based on feedback. This iterative process can incorporate external memory systems, enabling the agent to track historical actions and environmental states, further improving the decision-making process over time~\cite{zhang2024survey, hu2024rag}.

\subsection{From LLMs to LAMs}

LAMs build upon the foundational capabilities of LLMs but are specifically optimized for action-oriented tasks. They are designed to perform actions in both physical and digital environments, interpreting user intentions from various data forms, automating processes as per user requirements, planning for task completion, and interacting with the world~\cite{zeng2023large}. This evolution signifies a shift from passive language interaction to generating action sequences that are grounded in real-world contexts.

\begin{figure}[tb]
    \centering
    \includegraphics[width=\columnwidth]{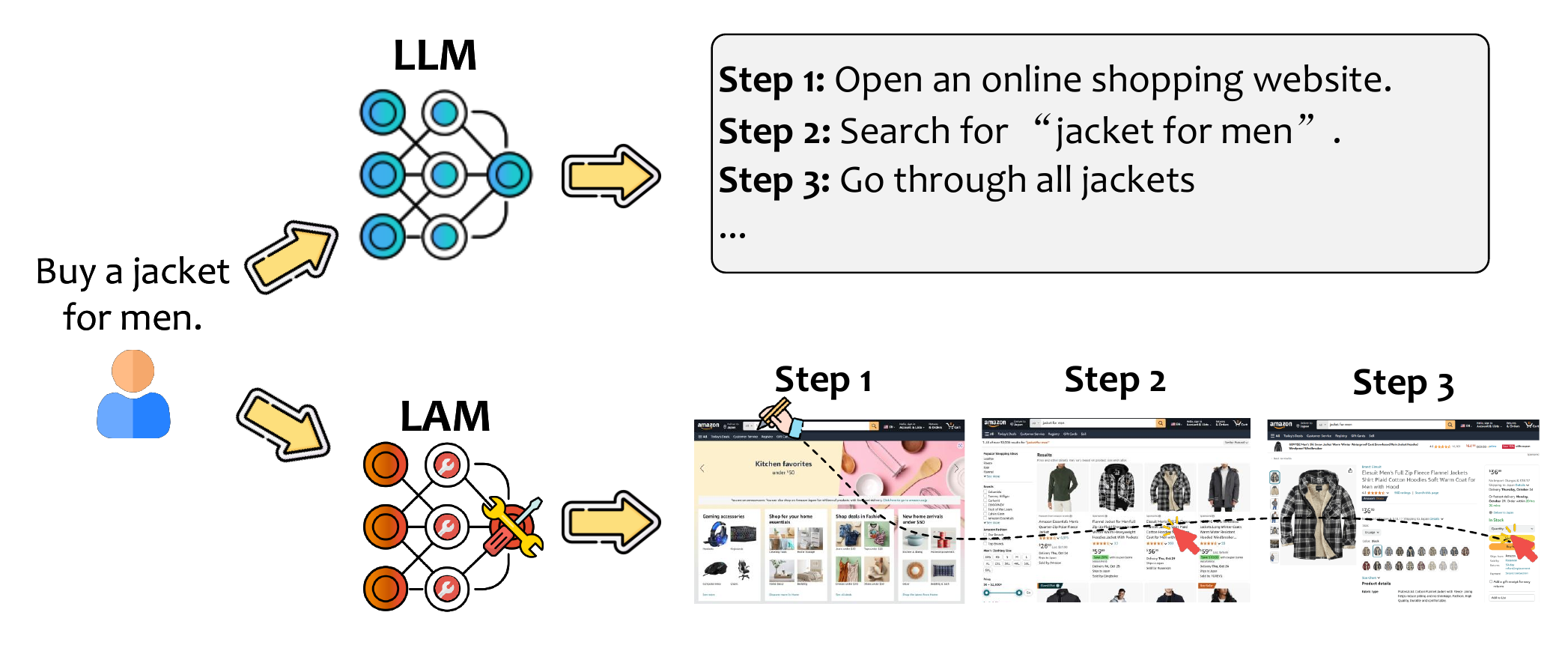}
    \vspace*{-3em}
    \caption{The objective difference between LLMs and LAMs.}
    \label{fig:output}
\end{figure}

An illustrative example is shown in Figure~\ref{fig:output}. An LLM can comprehend a user's request to purchase a jacket and generate a detailed textual plan or recommendation, but it cannot autonomously complete the transaction on a website. In contrast, a LAM leverages this foundational understanding to generate action sequences that directly interact with the website, completing the request on the user's behalf. This ability to transition from understanding to execution bridges the gap between the model and real-world applications, moving beyond mere language output to tangible outcomes.

Furthermore, due to their specialization in specific domains or tasks, LAMs can be smaller in scale compared to general-purpose LLMs while achieving comparable or superior performance within their operational scope. By focusing on a narrower range of tasks, LAMs prioritize efficiency and effectiveness, leveraging targeted data and optimized architectures to reduce computational overhead without sacrificing capability. This specialization not only makes LAMs more practical for deployment in real-world applications but also opens opportunities for developing lightweight models that can operate in resource-constrained environments.

The evolution from LLMs to LAMs is achieved through specialized training and integration with agent systems. These systems enable LAMs to translate their inferences into real-world actions, bridging the gap between understanding and execution. Thus, LAMs not only enhance the functionality of LLMs but also redefine their applicability in real-world scenarios.

\subsection{Key Characteristics of LAMs}
LAMs are distinguished by advanced capabilities that enable them to perform complex tasks effectively. These characteristics include:

\subsubsection{Interpretation of User Intentions}
A fundamental capability of LAMs is the ability to accurately interpret user intentions from diverse forms of input. These inputs may include natural language requests, voice commands, images, or videos, such as device screenshots or instructional videos~\cite{cheng2024seeclick}. User inputs are often abstract or implicit~\cite{chen2024large}, requiring LAMs to leverage their internal knowledge and complementary information to discern the true intent behind the input. This process involves understanding nuances, disambiguating instructions, and inferring unstated objectives. LAMs must translate these user intentions into actionable plans and steps, facilitating subsequent interactions with the environment to fulfill the user's objectives. This requires a robust foundation in LLMs, particularly those with multi-round conversational capabilities~\cite{shah2023using}, enhancing LAMs' proficiency in engaging with users to accurately understand and execute their requests.

\subsubsection{Action Generation}
The hallmark feature of LAMs is their capacity for action generation grounded in the environment. LAMs translate user intentions into actionable steps that can be executed within specific contexts. These actions can take various forms: operations on graphical user interface (GUI) elements, API calls for software applications, physical manipulations performed by robots, invoking other AI agents or models, or autonomously generating code or combining meta-actions~\cite{carta2023grounding}. By incorporating detailed knowledge of the environment, including available actions, system states, and expected inputs, LAMs can select appropriate actions and apply them correctly to meet user requests. This involves not only executing predefined actions but also adapting to new situations by generating novel action sequences when necessary.

\subsubsection{Dynamic Planning and Adaptation}
LAMs exhibit a sophisticated capability for dynamic planning and adaptation, which is crucial for handling complex user requests that span multiple steps~\cite{guan2023leveraging}. They can decompose a complex task into several subtasks, each further broken down into specific action steps. This hierarchical planning enables LAMs to approach task execution with a forward-looking perspective, anticipating future requirements and potential obstacles. Moreover, as the execution of each action alters the state of the environment, LAM will react to these changes, adapting and revising their plans and actions accordingly~\cite{shinn2024reflexion}. This flexibility ensures robustness in dynamic scenarios where deviations from initial expectations are common. For instance, if an unexpected error occurs or a resource becomes unavailable, a LAM can replan and adjust its actions to still achieve the desired outcome.

\subsubsection{Specialization and Efficiency}
LAMs are fine-tuned for executing specialized sequences of actions within specific environments~\cite{cheng2024seeclick}. By focusing on particular domains, LAMs achieve a high degree of accuracy and adaptability, outperforming general-purpose LLMs in targeted applications. This specialization allows LAMs to encode comprehensive knowledge about the environment deeply into their architecture, including available actions, system constraints, and contextual nuances. As a result, LAMs can operate more efficiently, reducing computational overhead and improving response times. Furthermore, since LAMs are expected to complete actionable tasks within a more limited scope, their scale can be smaller compared to general-purpose LLMs while achieving a comparable level of performance within that specific domain. This makes LAMs more practical for deployment in real-world applications, including resource-constrained environments such as edge devices or local systems.

\subsubsection{Summary}
In summary, LAMs transcend the basic functionality of converting user requests into a series of steps by comprehending the underlying logic that interconnects and contextualizes these actions. They understand sequence dependencies---why certain steps must precede or follow others---and recognize when to adapt the plan to accommodate changing circumstances. LAMs extend AI systems into the realm of actionable intelligence. This significantly enhances their ability to autonomously perform complex, real-world tasks, making them invaluable in applications requiring precise interaction and manipulation within defined operational contexts.

\begin{figure*}[t]
    \centering
    \includegraphics[width=0.8\textwidth]{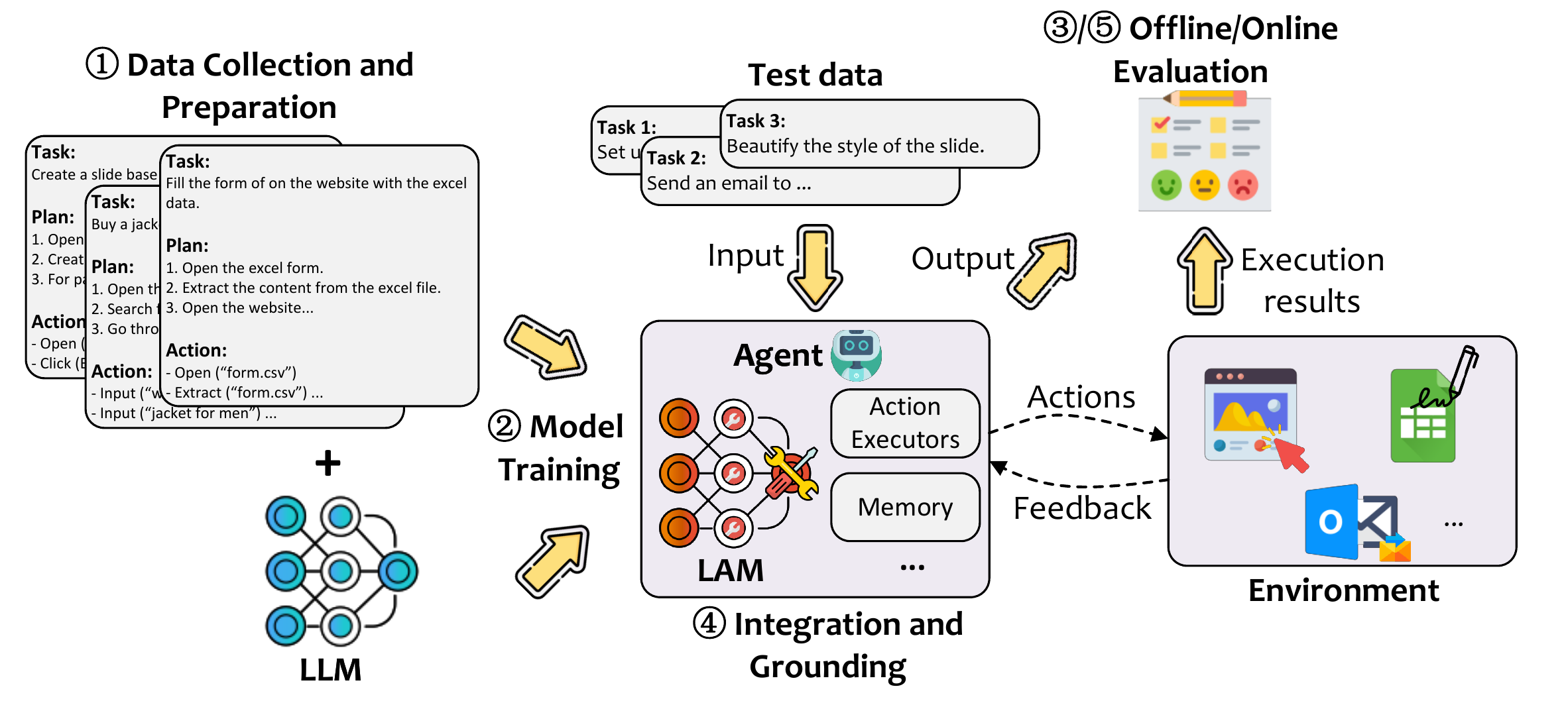}
    \vspace*{-1.5em}
    \caption{The process pipeline for LAM development and implementation.}
    \label{fig:workflow}
\end{figure*}

\subsection{From Inception to Implementation}

LAMs have the potential to significantly extend the impact of LLMs by enabling tangible interactions with real-world environments. To harness this potential, an LAM must be developed from the ground up and deployed within a real-world application, allowing it to operate effectively in a physical environment. This process involves 5 critical steps, as shown in Figure~\ref{fig:workflow}:
\begin{enumerate}
    \item \textbf{Data Collection and Preparation (Section~\ref{sec:data})}: The first step involves gathering and curating the necessary data for the specific use case. This includes not only user queries but also environmental context, potential actions, and any other relevant data required to train the LAM effectively. The data must undergo cleaning and pre-processing before it is used for training or fine-tuning a LAM.

    \item \textbf{Model Training (Section~\ref{sec:train})}: Using the prepared data, the next step is to train the LAM. This training process can involve various techniques such as supervised fine-tuning and reinforcement learning to ensure the model can perform the desired actions accurately and efficiently.

    \item \textbf{Offline Evaluation (Section~\ref{sec:offline_eva}):} After obtaining the LAM, we evaluate its performance using an offline dataset to verify its reliability in a controlled, static environment.

    \item \textbf{Integration and Grounding (Section~\ref{sec:grounding})}: The LAM is integrated into an agent framework that serves as its operational platform. This involves grounding the model with the ability to interact with external tools, maintain memory, and interface with the environment. By equipping the LAM with these capabilities, it becomes capable of making meaningful impacts in the physical world.

    \item \textbf{Online Evaluation (Section~\ref{sec:online_eva})}: Finally, the performance of the LAM must be rigorously evaluated in the real environment from multiple perspectives, including accuracy, efficiency, and effectiveness in completing tasks. This step is crucial to ensure that the LAM functions as intended and meets the desired operational standards.
\end{enumerate}
Through these steps, LAMs can be effectively developed and deployed to bring LLMs' capabilities into real-world applications, enabling them to interact with and manipulate the physical environment, thereby making a tangible impact.

In the following sections, we use the Windows GUI agent UFO~\cite{ufo}\footnote{\url{https://github.com/microsoft/UFO}} as a case study to illustrate the process of building a robust LAM from the ground up. This LAM will serve as the core inference engine for UFO, enabling it to autonomously fulfill user requests within the Windows OS environment. While this example focuses on a Windows GUI agent, the outlined steps can be adapted for developing LAMs in other scenarios or for different applications.

\section{Data Collection and Preparation\label{sec:data}}

\begin{figure*}[t]
    \centering
    \includegraphics[width=0.8\textwidth]{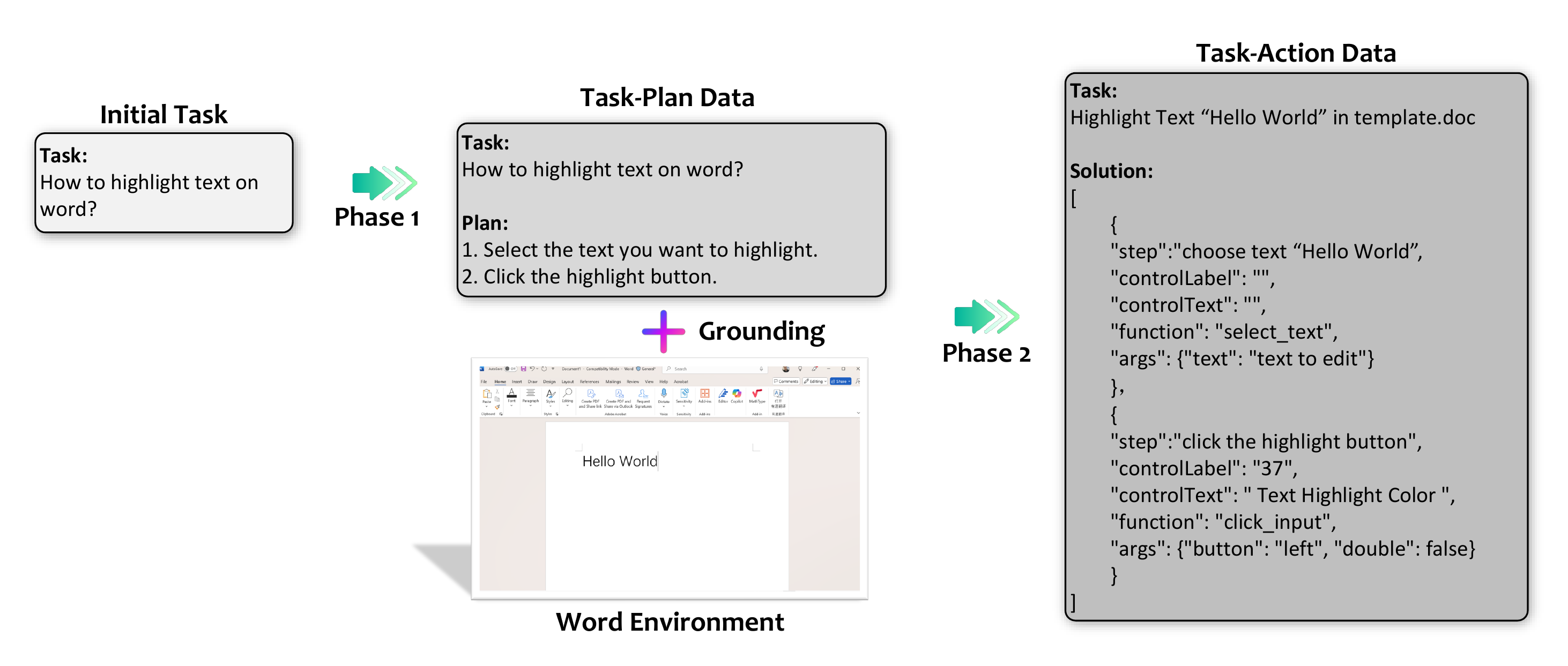}
    \vspace*{-1.5em}
    \caption{The two-phrase data collection and preparation process.}
    \label{fig:data}
\end{figure*}
Data is a cornerstone in training LLMs, where high-quality data significantly enhances their performance~\cite{wang2023data,li2024effects}. Similarly, LAMs require well-prepared, high-quality action-oriented data during the supervised fine-tuning phase. Off-the-shelf LLMs often face challenges when interacting with real-world environments. These difficulties typically arise from either a lack of domain-specific knowledge or the generation of hallucinated outputs that fail to be actionable. To mitigate these issues, we adopt a two-phase data collection approach: \textit{task-plan collection} and \textit{task-action collection}, as shown in Figure~\ref{fig:data}. Specifically:

\begin{enumerate} 
    \item \textbf{Task-Plan Data Collection:}
    In this phase, we collect data consisting of tasks and their corresponding plans. Tasks are user requests expressed in natural language, while plans are detailed, step-by-step procedures designed to fulfill these requests. For example, a task such as \textit{``How to change the font size in Word?''} would have a corresponding plan outlining the steps required to complete the task. This data is used to fine-tune the model to generate effective plans and improve its high-level reasoning and planning capabilities. However, task-plan data cannot be directly executed in the environment, requiring the following data conversion phase.
    \item \textbf{Task-Action Data Collection:} In this phase, the task-plan data is converted into task-action data, which includes tasks, plans, and the associated action sequences needed to execute those plans. Tasks and plans are refined to become more concrete and grounded within a specific environment. Action sequences are generated at this stage, such as \texttt{select\_text(\\text="hello")} or \texttt{click(on=Button("20"), how="left", double=False)}, which represent actionable instructions capable of directly interacting with the environment. This enriched data provides the necessary granularity for training an LAM to perform reliable and accurate task executions in real-world scenarios.
\end{enumerate}

The task-plan data aims at enhancing the model's high-level planning capabilities, allowing it to generate detailed, step-by-step plans based on user requests. Meanwhile, the task-action data focuses on refining the model's ability to execute these plans by converting each planned step into a concrete, executable step or sequence while considering environmental feedback. The data collection and preparation pipeline ensures that the model is capable of both high-level planning and low-level action execution, thereby bridging the gap between natural language plans and executable actions.

In the following sections, we detail the methodologies employed for data collection, pre-processing, and integration of task-plan and task-action data. We illustrate how these steps enable the LLM to LAM transformation.

\subsection{Task-Plan Data\label{sec:task_plan}}
\begin{figure}[t]
    \centering
    \includegraphics[width=\columnwidth]{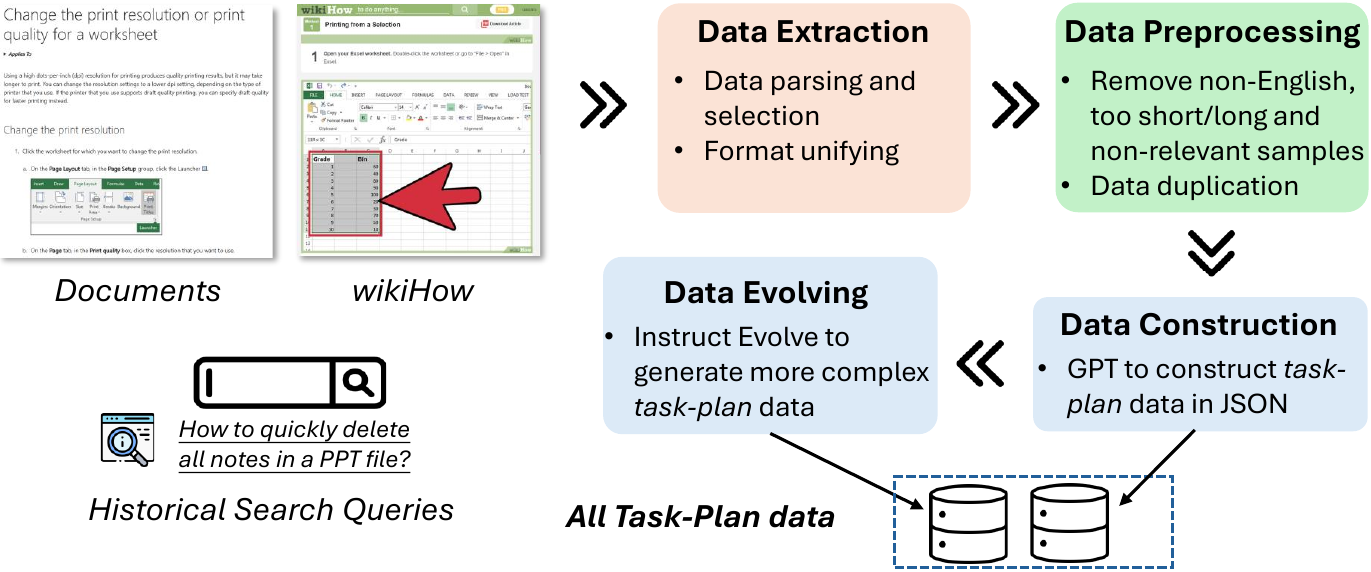}
    \caption{The pipeline to construct the task plan data.}
    \label{fig:task_plan_data_pipeline}
\end{figure}

Figure~\ref{fig:task_plan_data_pipeline} outlines a multi-step pipeline for collecting and processing task-plan data, essential for training LAMs. The process begins with gathering raw data from diverse sources, including application documentation, WikiHow, and historical search queries. This is followed by structured pre-processing to ensure that the data is high-quality and relevant to specific tasks.

\subsubsection{Data Sources}

\begin{enumerate}[leftmargin=*]
    \item \textbf{Application Documentation:}
    Documentation and usage manuals for software applications provide authoritative task descriptions. These resources, maintained by product teams, are considered highly reliable. Relevant documentation, such as M365 documentation\footnote{\url{https://learn.microsoft.com/en-us/microsoft-365/?view=o365-worldwide}}, is crawled, with outdated or inaccessible pages being filtered out. The HTML content is converted into markdown format, and GPT-4o is used to extract task-plan pairs in the desired structured format.
    
    \item \textbf{WikiHow:}  
    WikiHow\footnote{\url{https://www.wikihow.com/Main-Page}} hosts a wide range of how-to articles, including application-specific operational guides. Webpages related to Windows platform applications are crawled, and GPT-4o extracts task and plan components, ensuring the resulting data aligns with the desired structured format.
    
    \item \textbf{Historical Search Queries:}  
    Search engine logs provide insight into real user demands, addressing gaps not covered by formal documentation. From Bing search logs, a 1\% sample of queries mentioning application names (\eg Word, Excel, PowerPoint) from the past year was taken. 
    
\end{enumerate}

\subsubsection{Data Extraction and Pre-Processing}

The initial step in processing raw data involves parsing to extract task-relevant content while filtering out unnecessary or irrelevant information. This includes removing non-English entries, samples that are excessively short or long based on predefined heuristics, and data unrelated to actionable tasks (\eg content focused on smartphone operations). The filtered data is then standardized into a unified format for further processing.

\subsubsection{Data Construction}

To create structured JSON samples, GPT-4o is employed to extract and format tasks along with their associated plans. For historical search queries, synthetic data is generated to enrich the raw input, addressing the common issue of insufficient context. GPT-4o reformulates these queries into complete, sentence-like user requests, ensuring consistency across all data sources and facilitating effective downstream processing.

The resulting dataset contains structured JSON samples, with each entry including a unique task identifier (task\_id), the task description (task), and a step-by-step plan (plan). An example is shown below:

\begin{lstlisting}[language=json,firstnumber=1]
{"task_id": "word_032",
 "task": "Add a border to a page in Word",
 "plan": [
    1. Go to Design > Page Borders.  
    2. Make selections for how you want the border to look. 
    3. To adjust the distance between the border and the edge of the page, select Options. Make your changes and select OK.
    4. Select OK.
    ]
}
\end{lstlisting}
With the above process, we initially collected a total of 29,182 task-plan data samples.

\subsubsection{Data Evolving}
\label{sec:data_evolving}
With the initial dataset processed, we employ data augmentation techniques to enhance its diversity and complexity. Inspired by WizardLM~\cite{xu2023wizardlm} and AgentGen~\cite{r_agentgen}, we use GPT-4o to evolve the raw task to generate new task-plan pairs, improving the model's ability to follow instructions and handle more complex tasks.

The data evolving process generates new tasks from the original ones by introducing additional complexity, constraints, or steps while preserving relevance. The guidelines for task evolution are as follows: \begin{itemize} 
\item[--] The evolved task must be executable step-by-step on a Windows OS or application. 
\item[--] The evolved task should include additional requirements, increasing its complexity without exceeding 20 extra words. \item[--] The evolved task must remain concise and related to the original task. 
\end{itemize}

For each evolved task, GPT-4o generates a corresponding plan adhering to the following guidelines: \begin{itemize} 
\item[--] The plan must provide correct and actionable steps for Windows environments or applications. 
\item[--] The plan should be concise and highlight critical action objects using bold emphasis. 
\end{itemize}

This augmentation process results in a richer dataset where tasks become progressively more challenging, and plans incorporate domain-specific knowledge. For example:

\textit{Raw task}: Create a drop-down list in Excel for Office 365. 

\textit{Evolved Task}: Create a dependent drop-down list in Excel for Office 365, where selecting an item from the first list filters options in the second list. 

\textit{Evolved Plan}: 

\begin{itemize}
    \item[--] Prepare your data by organizing it into two columns. The first column contains items for the primary drop-down list, and the second column contains items for the dependent list.
    \item[--] Name your ranges for the first and second lists. 
    \item[--] Create the primary drop-down list using Data Validation.
    \item[--] Use the INDIRECT function to create the dependent drop-down list linked to the first selection. 
    \item[--] $\cdots$
\end{itemize}

Using data augmentation, we increased the original task-plan dataset by 150\%, generating a larger pool of samples. This augmentation significantly enhances the diversity and complexity of the dataset, allowing the model to learn from a broader range of scenarios and develop robust planning capabilities. The augmented data introduces more challenging tasks and detailed plans, further enriching the training process and enabling the LAM to handle complex real-world applications effectively.

\subsection{Task-Action Data  \label{sec:task_action}}

The task-plan data collected in the previous stage provides high-level, step-by-step plans for resolving user-requested tasks, serving as general guidelines. However, these plans are textual and not directly executable in a real-world environment. For instance, a task-plan data sample for the task ``Highlight text in document'' outlines the necessary steps but does not translate into actionable instructions for interacting with the application's GUI. This gap highlights the need for actionable task-action data to bridge the divide between planning and execution. To enable LAMs to produce actionable outputs, we generate task-action data derived from the previously collected task-plan data. Task-action data captures the granular interactions required to complete a task in the application environment, including GUI navigation, button clicks, and responding to environmental feedback.

\begin{figure}[htbp]
    \centering
    \includegraphics[width=\columnwidth]{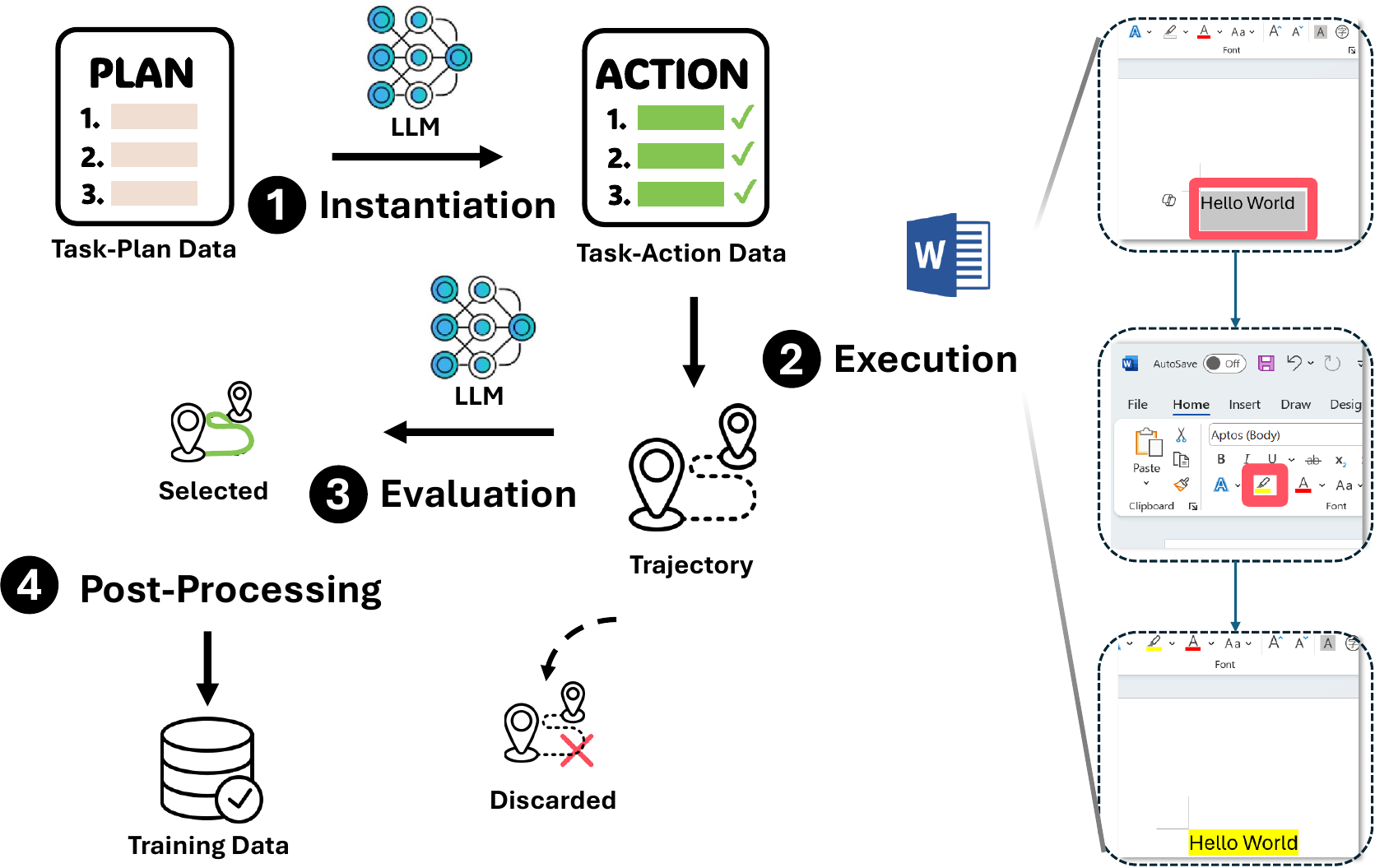}
    \caption{The pipeline of task-action data conversion and collection.}
    \label{fig:action-pipeline}
\end{figure}

Traditional approaches for action data collection often involve manual or agent-based annotation for each task, which is both costly and labor-intensive. To address these limitations, we propose an efficient, fully automated, and low-cost pipeline that leverages LLMs and real-world application interactions. This pipeline consists of four stages, as depicted in Figure~\ref{fig:action-pipeline}: \textbf{Instantiation}, \textbf{Execution}, \textbf{Evaluation}, and \textbf{Post-Processing}. Specifically,

\begin{enumerate} 
    \item \textbf{Instantiation:}
    In this stage, the task-plan data is transformed into an executable trajectory. Using an LLM, each task is instantiated with specific operational objects, and related high-level plan is instantiated into a concrete sequence of actions that can be directly executed in the application environment. 
    
    \item \textbf{Execution:} 
    The instantiated trajectory is then executed within the real-world application environment. During this stage, the system interacts with the application's GUI to carry out the specified actions. For example, the instantiated trajectory for highlighting text would involve selecting the appropriate text, navigating to the highlight tool, and applying the highlight. The result of this execution is the captured executed trajectory, including any feedback or environmental changes observed during the process.
    
    \item \textbf{Evaluation:} 
    Once the execution is complete, the trajectory is evaluated for correctness using an LLM. The evaluation stage verifies whether the executed trajectory successfully accomplishes the intended task. This involves comparing the observed outcomes with the expected results outlined in the task-plan data. Tasks that fail to meet the criteria are flagged for review, while successful executions are retained for further processing.
    
    \item \textbf{Post-Processing:} 
    In the final stage, successful task-action trajectories undergo post-processing to ensure consistency, completeness, and readiness for training. This includes refining the data format, ensuring compatibility with the training pipeline, and annotating the data with relevant metadata (\eg task IDs, execution time, and step-by-step feedback). The post-processed task-action data is then added to the training dataset, enabling the LAM to learn from real-world interactions.
\end{enumerate}
The pipeline minimizes human intervention and reduces the number of LLM calls required, significantly improving scalability and efficiency.

\subsubsection{Instantiation}
The task-plan data are primarily collected from help documents or public websites, creating a gap between the generalized task-plan data and the specific requirements needed for execution within a particular environment. A common issue is the lack of specificity. For instance, the task \textit{``highlight text in document''} does not specify actionable objects, such as \textit{``which text''} or \textit{``which document''}. This lack of detail poses significant challenges in executing tasks within real-world applications.

To address this problem, we instantiate the task-plan data to impute target objects and related functions. First, we prepare template Word files to serve as specific targets for the actions. These template files include various Word components such as paragraphs, tables, and figures. Each template file is accompanied by a description indicating its content, providing context for grounding actions. Several sample template files can be found in Appendix~\ref{append:templates-word}.

Given a task-plan data sample, the task description is matched with the template file descriptions to select an appropriate template file as the target for actions. GPT-4 is then prompted to instantiate the task-plan with target objects present in the selected template file (detailed prompts can be found in Appendix~\ref{append:instance}). Simultaneously, we filter relevant functions from the available function pool using the task description, allowing the instantiation process to populate the task-action data with specific functions and their input parameters. 

\begin{figure}[htbp]
    \centering
    \includegraphics[width=\columnwidth]{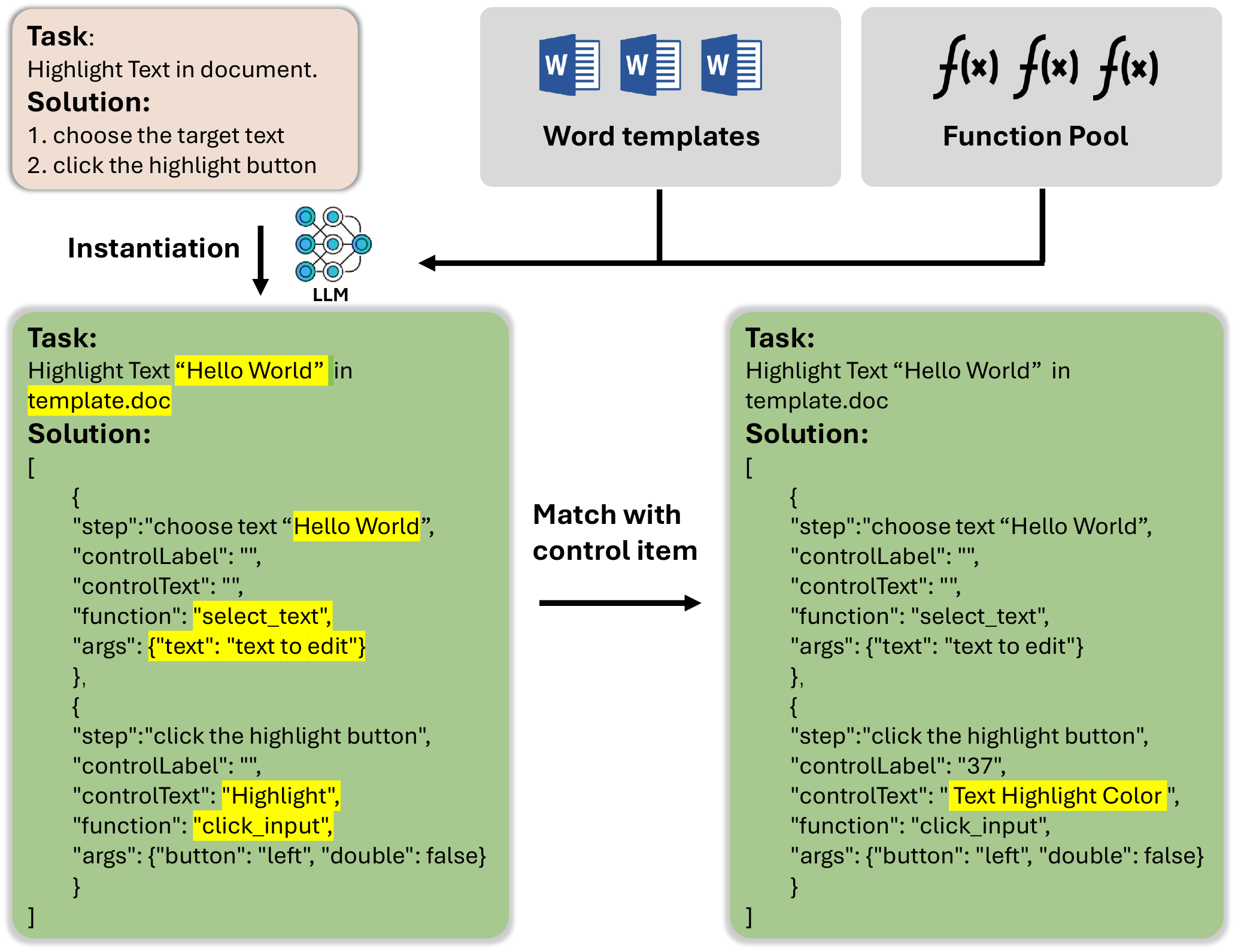}
    \caption{An example of task instantiation.}
    \label{fig:instantiate}
    \vspace{-5mm}
\end{figure}

As a result of this process, the task description becomes more concrete and grounded in a specific environment, while the corresponding action sequences needed to complete the task are generated. Figure~\ref{fig:instantiate} provides an example of the instantiation process. Notably, the task-action data is not directly generated with GPT-4 due to the risk of hallucinations. Instead, instantiating grounded task-plan data ensures the generation of more reliable and faithful step-by-step actions.

\subsubsection{Execution\label{sec:offline}}
To ensure that the steps in the instantiated task-plan data are accurate and truly actionable, the execution stage verifies the action sequence by matching control items with the real application environment and performing the specified actions. This process validates the task-action data, ensuring its correctness and compatibility with the application GUI.

For instance, as shown in Figure~\ref{fig:instantiate}, the control item \textit{``Text Highlight Color''} with its associated control label is retrieved using the action text \textit{``Highlight''} from the control item pool. The corresponding task-action data is then executed in the application without further intervention from the LLM. During execution, if an error occurs (\eg a mismatch between the predicted control item and the actual environment), the instantiated task is discarded. Conversely, if all actions in the task execute successfully, the action-validated task is forwarded to the evaluation stage described in the following section. Additionally, screenshots of the application environment are captured after each step in the execution process, forming a detailed trajectory to assist in subsequent evaluation.

It is important to note that the instantiated task-action data is not guaranteed to be valid. Since the data is generated through a single GPT-4 call based on task-plan data, it lacks the step-by-step refinement that might be necessary for certain tasks. In some cases, execution results from previous steps are required to instantiate subsequent steps accurately. In such scenarios, the one-call instantiated task-action data may fail in validation and is removed from the dataset. This execution stage bridges the gap between planning and action, ensuring that task-action data is actionable, robust, and aligned with real-world application requirements.

\subsubsection{Evaluation}\label{sec:evaluate}
Even if the task-action data is successfully executed in the real application without errors, further evaluation is required to ensure its validity. Some tasks may be incorrectly instantiated from the task-plan data, resulting in trajectories that, while executable, do not fulfill the original task description. Similarly, the executed results might fail to align with the intended task outcomes. For evaluation, we utilize the instantiated task along with its execution trajectory, which includes:
\begin{itemize}
    \item[--] Consecutive actions performed during execution.
    \item[--] Screenshots captured before and after each action.
    \item[--] Environmental changes observed between the initial and final states\footnote{More specifically, we compare the \texttt{.xml} files which is the underlying data representation of Microsoft Word.}.
\end{itemize}
Using this comprehensive trajectory, we prompt GPT-4o to evaluate whether the executed task aligns with the original task description and achieves successful completion. The evaluation considers both the sequence of actions and the resulting application state. The process assigns a "task-complete" key to indicate the outcome as "yes," "no," or "unsure." If the task is evaluated as \texttt{"yes"}, the trajectory is deemed successful; otherwise, it is classified as a failure. The detailed prompt used for this evaluation is provided in Appendix~\ref{append:evaluation}. This evaluation step ensures that only valid, accurate task-action data is included in the training dataset, contributing to the reliability and robustness of the LAM.

\subsubsection{Post-Processing}
As noted in Section~\ref{sec:offline}, a trajectory was recorded during the execution process. This trajectory includes:
\begin{itemize}
    \item[--] Screenshots captured at each step.
    \item[--] Environment states before and after each action.
    \item[--] Plans and corresponding actions for every step.
\end{itemize}
During the post-processing stage, these trajectories are combined with the original task requests to generate synthetic step-wise training data. The resulting data format uses the task request as input and LAM's plan and actions as output. This structured format is critical for training LAMs to map task requests to actionable sequences effectively. The detailed template for the data format can be found in Appendix~\ref{sec:template}.

\begin{figure*}[t]
    \centering
    \includegraphics[width=1.\textwidth]{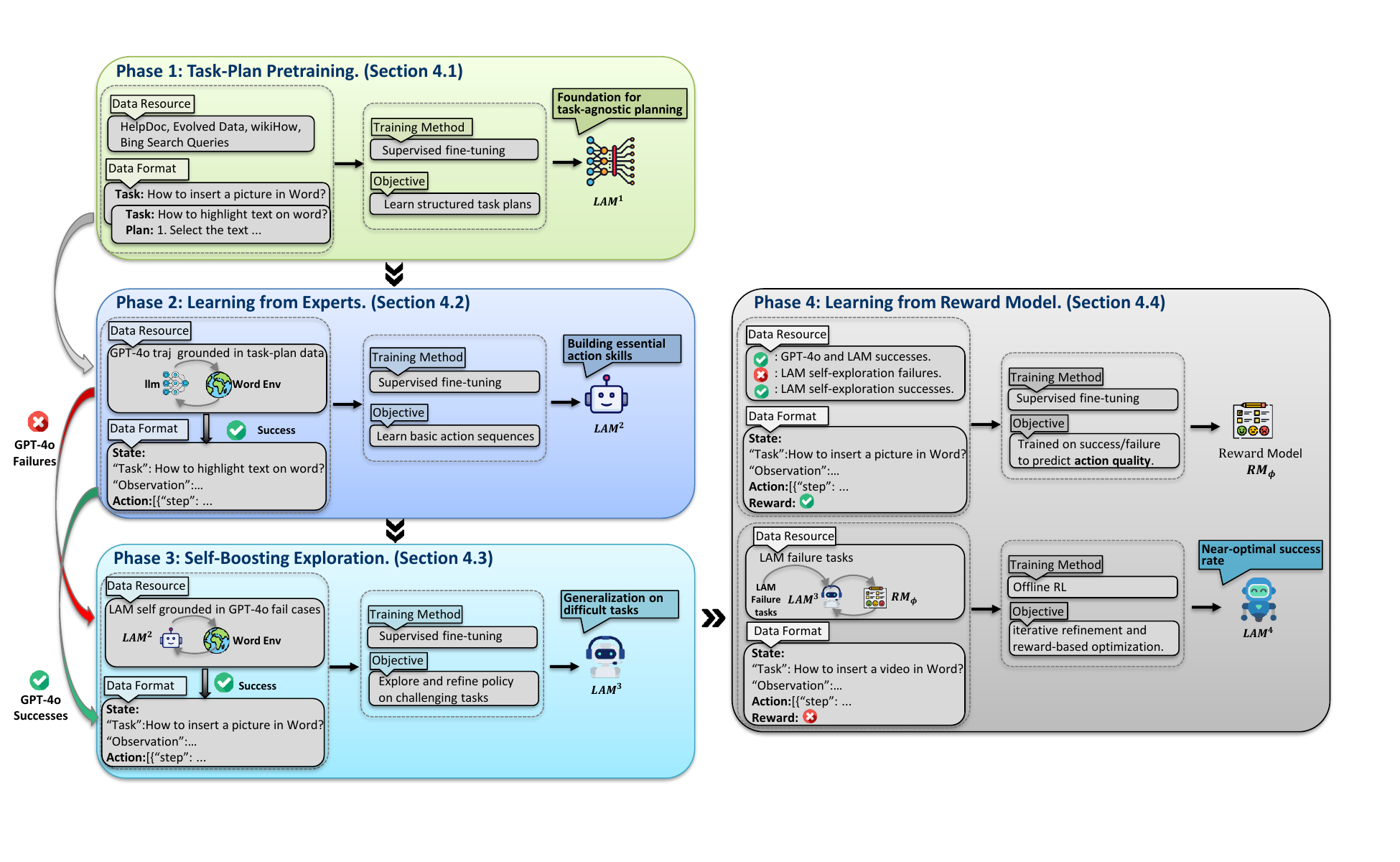}
    \vspace{-5em}
    \caption{The overview of LAM training pipeline.
    }
    \label{fig:LAM_framwork}
\end{figure*}

\section{Model Training\label{sec:train}}

Our objective is to develop an LAM from scratch that can map user inputs to appropriate plans and executable actions, ultimately enabling complex task completion. To achieve this, we adopt a staged training strategy consisting of four phases, each building upon the previous one. As illustrated in Figure~\ref{fig:LAM_framwork}, these phases guide the model from learning structured task plans, to imitating expert demonstrations, to self-boosting from its own successes, and finally leveraging reward-based optimization. Throughout these stages, the model progressively evolves from \(\text{LAM}^1\) to \(\text{LAM}^4\).

At a high level, \textbf{Phase 1: Task-Plan Pretraining} provides a strong foundation by teaching the model to generate coherent, step-by-step plans for various tasks. \textbf{Phase 2: Learning from Experts} then introduces action trajectories labeled by GPT-4o, enabling \(\text{LAM}^2\) to align its plan generation with actionable steps. However, relying solely on expert successes limits diversity and adaptability. To address this, \textbf{Phase 3: Self-Boosting Exploration} encourages the model to tackle tasks that even GPT-4o failed to solve, autonomously generating new success cases and evolving into \(\text{LAM}^3\). Finally, \textbf{Phase 4: Learning from a Reward Model} incorporates reinforcement learning (RL) principles, allowing \(\text{LAM}^4\) to learn from both successes and failures, refining its decision-making in complex, previously unseen scenarios.
Table~\ref{tab:training_data_summary} summarizes the data used in each phase. Each phase uses different training objectives, namely \textit{(i)} task-plan pretraining (phase 1) and \textit{(ii)} decision-making training (phase 2-4), as detailed in Appendix~\ref{app:obj}.

\begin{table*}[htb]
\small{
\centering
\caption{Training data summary for each phase of LAM training.}
\begin{tabular}{c|l|l|l|c}
\hline
\textbf{Model} & \textbf{Data Type}                & \textbf{Data Source}                 & \textbf{Input \(\to\) Output Format}                          & \textbf{Data Size} \\ \hline
\( \text{LAM}^1 \) & Task-Plan Pairs                & \makecell{Application documentation, WikiHow, \\historical search queries, evolved data}   & \( t_i \to P_i \)                             & 76,672 tasks       \\ \hline
\( \text{LAM}^2 \) & Task-Action Trajectories   & GPT-4o                          &\( s_t \to a_t \)                  & 2,192 trajectories \\ \hline
\( \text{LAM}^3 \) & Task-Action Trajectories & \( \text{LAM}^2 \) + GPT-4o       & \( s_t \to a_t \)    & 2,688 trajectories   \\ \hline
\( \text{LAM}^4 \) &  Task-Action-Reward Trajectories  & RM + \(\text{LAM}^3\)          & \( (s_t, r_t) \to a_t \)            & 1,788 trajectories \\ \hline
\( \text{Reward Model} \) &  Task-Action-Reward Trajectories  & GPT-4o + \(\text{LAM}^3\)          & \( (s_t, a_t) \to r_t \)             & 4,476 trajectories \\ \hline
\end{tabular}
\label{tab:training_data_summary}
}
\end{table*}

\subsection{Phase 1: Task-Plan Pretraining}

The initial stage focuses on imparting a broad understanding of how tasks can be decomposed into logical steps. We start with Mistral-7B~\cite{jiang2023mistral} as the base model. A total of {\bf 76,672} task-plan pairs \((t_i, P_i)\) are collected from various sources, including application help documentation, WikiHow, and historical search queries. Of these, 29,182 pairs are sourced directly, while 47,490 are generated via data evolution techniques (as described in Section~\ref{sec:data_evolving}), enriching the dataset with more complex and diverse tasks.

In this phase, \(\text{LAM}^1\) is trained via supervised fine-tuning (SFT) to predict the correct plan sequence \(P_i\) for a given task \(t_i\):
\[
\mathcal{L}_{\text{SFT}}(\text{LAM}_\theta^1) = \frac{1}{N} \sum_{i=1}^N \mathcal{L}_{\text{CE}}(P_i^{\text{pred}}, P_i^{\text{true}}).
\]
Here, \(\mathcal{L}_{\text{CE}}\) denotes the cross-entropy loss, and \(N\) is the number of tasks. Although no actions are generated at this stage, \(\text{LAM}^1\) gains a robust task-agnostic planning capability. This knowledge will prove critical in guiding the model’s action execution in later phases, ensuring that the agent understands the logical structure of tasks before attempting to perform them.

\subsection{Phase 2: Learning from Experts}

While \(\text{LAM}^1\) can produce structured plans, it lacks the ability to execute them. In Phase 2, we introduce expert-labeled task-action trajectories from GPT-4o~(Section~\ref{sec:task_action}) to teach the model how to perform actions. The illustrative application in this paper is the Microsoft Word environment, where we have {\bf 2,192} successful expert trajectories. Each trajectory consists of a sequence of state-action pairs \((s_t, a_t)\), representing observed UI states and the corresponding actions to progress the task.

We split these 2,192 trajectories into a training set of 1,757 and a test set of 435 trajectories, providing a total of 3,959 steps for training. By imitation learning \(\text{LAM}^1\) on these successful action sequences, we obtain \(\text{LAM}^2\). The objective is to minimize:
\[
\mathcal{L}_{\text{SFT}}(\text{LAM}_\theta^2) = \frac{1}{N} \sum_{i=1}^N \sum_{t=1}^{T_i} \mathcal{L}_{\text{CE}}(\text{LAM}_\theta^2(s_t), a_t),
\]
where \(N\) is the number of trajectories and \(T_i\) is the number of steps in trajectory $i$. By imitating the expert’s policy, \(\text{LAM}^2\) transforms from a passive planner into a model capable of executing actions aligned with its plans, grounding its reasoning in the real application environment.

\subsection{Phase 3: Self-Boosting Exploration}

Up to Phase 2, \(\text{LAM}^2\) only learns from successful trajectories provided by GPT-4o. This limits diversity and adaptability, as the model never sees how to handle situations that even GPT-4o could not deal with. To overcome this limitation, Phase 3 introduces self-boosting exploration.

Here, we revisit failed GPT-4o trajectories, \ie tasks that GPT-4o did not complete successfully, and let \(\text{LAM}^2\) attempt them. Using the ReAct mechanism~\cite{yao2022react, shinn2024reflexion}, \(\text{LAM}^2\) interacts with the environment and tries alternative strategies for these challenging tasks. From these attempts, we sampled 2284 GPT-4o failed tasks and then collect {\bf 496} newly successful trajectories generated by \(\text{LAM}^2\) itself. These self-labeled successes, combined with the original 2,192 GPT-4o successes, form an augmented dataset.

We then fine-tune \(\text{LAM}^2\) on this enriched data, yielding \(\text{LAM}^3\):
\[
\mathcal{L}_{\text{SFT}}(\text{LAM}_\theta^3) = \frac{1}{N} \sum_{i=1}^N \sum_{t=1}^{T_i} \mathcal{L}_{\text{CE}}(\text{LAM}_\theta^3(s_t), a_t).
\]
This self-boosting step allows the model to learn from its own newly discovered solutions, overcoming previous limitations and improving adaptability. By leveraging planning knowledge from Phase 1 and expert strategies from Phase 2, \(\text{LAM}^3\) becomes more resourceful, even in scenarios with sparse or absent expert guidance.

\subsection{Phase 4: Learning from a Reward Model}
Despite the improvements, Phases 1–3 focus on successes or expert-like behavior. They offer limited insights into intermediate decision quality and fail to exploit learning opportunities presented by failed attempts. In Phase 4, we integrate reinforcement learning (RL) to address these shortcomings.

To this end, we design a two-stage approach, where we first The reward model (RM) is built using LAM$^3$
as the base model, with an additional output layer added to produce scalar values representing the quality of actions. Using the trained RM, we fine-tune \( \text{LAM}^4 \) in an offline RL setting. Here, the model refines its policy without additional environmental interactions, leveraging previously collected trajectories to learn from failures and improve action selection.

\subsubsection{Reward Model Training}

First, we train a reward model (RM) on both LAM$^3$'s successful (496) and failed (1788) trajectories and GPT-4o's successful trajectories (2192) gathered in previous phases. All steps in successful trajectories are assigned a reward of \(+1\), and all steps in failed trajectories a reward of \(-1\). This uniform, binary labeling of outcomes ensures the RM consistently captures overall trajectory quality. Formally:
\[
r_t = \text{RM}(s_t, a_t; \phi),
\]
where \(\phi\) presents the RM parameters, and \(r_t \in \{+1, -1\}\) is the assigned reward. The RM is trained via mean squared error (MSE) to approximate these ground-truth rewards.

The training dataset for the RM includes both failed and successful task-action trajectories generated by \(\text{LAM}^3\), as well as the successful trajectories from the collected task-action data. All steps in successful trajectories receive a reward of \(+1\), while every step in failed trajectories is assigned a reward of \(-1\). This uniform labeling strategy ensures that the RM consistently reflects overall trajectory quality and effectively guides policy optimization.


\subsubsection{Optimizing with Offline PPO \cite{schulman2017proximal}}
Armed with the RM to evaluate intermediate actions, we fine-tune \(\text{LAM}^4\) via offline PPO. This stage focuses on the 1,788 failure trajectories collected during Phase 3, providing a unique opportunity to learn from mistakes. The training objective of PPO  is:
\[
\begin{aligned}
\mathcal{L}_{\text{PPO}}(\text{LAM}^4_\theta) = 
\frac{1}{N} \sum_{i=1}^N \sum_{t=1}^{T_i} \min&\Bigg(
\frac{\text{LAM}^4_\theta(a_t | s_t)}{\text{LAM}^4_{\theta_{\text{old}}}(a_t | s_t)} \hat{A}_t, \\
&\text{clip}\Bigl(\frac{\text{LAM}^4_\theta(a_t | s_t)}{\text{LAM}^4_{\theta_{\text{old}}}(a_t | s_t)}, 1-\epsilon, 1+\epsilon\Bigr) \hat{A}_t
\Bigg),
\end{aligned}
\]
where \(\hat{A}_t\) denotes the advantage derived from RM-generated rewards, and \(\epsilon\) is a clipping parameter to ensure stable updates.

By incorporating signals from both successes and failures, \(\text{LAM}^4\) gains a deeper understanding of action quality. This RL-based fine-tuning helps the model generalize to complex, previously unseen scenarios, ensuring more robust and reliable decision-making.

\subsection{Summary}
The four-phase training pipeline incrementally builds a fully capable LAM. Phase 1 imparts a fundamental planning ability, Phase 2 incorporates expert knowledge for action execution, Phase 3 empowers the model to generate and learn from new successes, and Phase 4 leverages rewards from both successes and failures to optimize decision-making. By combining static knowledge with expert demonstrations, self-guided exploration, and reward-based refinement, we transform a general-purpose language model into a versatile LAM. This progressive training strategy ensures a robust, adaptive model ready to handle diverse and complex tasks.

\section{Offline Evaluations\label{sec:offline_eva}}
The offline evaluation results of \textbf{Task-Plan Pretraining Results (Phase 1)} and \textbf{Task-Action Results (Phases 2–4)} will be presented in this section. Offline evaluation allows us to systematically assess the performance of \( \text{LAM}^1 \) and subsequent phases (\( \text{LAM}^2 \), \( \text{LAM}^3 \), and \( \text{LAM}^4 \)) without interacting with the environment. This setup effectively provides a controlled and reproducible framework for comparing task success rates, precision, and recall metrics across models.

\subsection{Experiment Setup}
\subsubsection{SFT Training (Phase 1, 2, 3).}  
For supervised fine-tuning (SFT), the learning rate is set to $2 \times 10^{-5}$ with cosine decay and 2 warmup steps. The batch size is 16, and the training is conducted for 3 epochs on the training data. Loss is calculated only for the target tokens rather than the full input sequence, optimizing the efficiency of the fine-tuning process. The training is performed on 8 $\times$ A100 80G NVIDIA GPUs.

\subsubsection{Reward Training (Phase 4).}  
Reward scores are normalized to the range [0, 1] using sigmoid function. We employ the LoRA (Low-Rank Adaptation) method~\cite{hu2021lora} to train the reward model (RM). The LoRA parameters include rank of 8, LoRA alpha of 32, and LoRA dropout of 0.1. The task type is sequence classification. The training process uses learning rate of $2 \times 10^{-5}$ with linear decay, optimized with the AdamW optimizer, and spans 2 epochs. The training is conducted on 8 $\times$ A100 80G NVIDIA GPUs.

\subsubsection{PPO Training (Phase 4).}  
For Proximal Policy Optimization (PPO) training, we use a learning rate of $1.4 \times 10^{-5}$ and set the generated sample length to 256. The batch size is 8, and the mini-batch size is 1, with 4 PPO epochs and 1 gradient accumulation step per iteration. The target KL divergence is set to 0.1, and the initial KL coefficient is set to 0.2. To ensure robust training, reward values are normalized to the range [-0.5, 0.5]. The training is conducted on 8 NVIDIA A100 80G GPUs.

\subsection{Task-Plan Pretraining Results (Phase 1)}

\subsubsection{Evaluation Metrics.}
We evaluate \( \text{LAM}^1 \) on its ability to generate task plans. We use three metrics for this evaluation: \emph{(i)} \textbf{Task Success Rate (TSR)}, measuring whether the predicted plan matches the ground truth at the task level; \emph{(ii)} \textbf{Step Precision}, evaluating the proportion of predicted plan steps that appear in the ground truth; and \emph{(iii)} \textbf{Step Recall}, assessing the proportion of ground truth plan steps that are correctly predicted.

To compute these metrics, we leverage GPT-4o to compare each step of the \(\text{LAM}^1\) output with the corresponding ground truth steps. The counts of matched steps are then used to calculate the final evaluation metrics. Detailed prompt information for the evaluation can be found in Appendix~\ref{app:task-plan-eval}.

\begin{table}[t]
\centering
\small
\caption{Performance (\%) comparison of different models on planning.}
\begin{tabular}{l|c|c|c}
\hline
\textbf{Model} & \textbf{TSR (\%)} & \textbf{Step Precision (\%)} & \textbf{Step Recall (\%)} \\ \hline
$\text{LAM}^1$ & 82.2 & \textbf{54.7} & 55.7 \\ \hline
GPT-4o & \textbf{84.5} & 28.2 & \textbf{66.1} \\ \hline
Mistral-7B & 0.0 & 0.1 & 0.5 \\ \hline
\end{tabular}
\label{tab:task-plan-performance}
\end{table}

\begin{table*}[t]
\centering
\caption{Offline performance comparison across different models and metrics on decision making.}
\begin{tabular}{l|c|c|c|c|c|c}
\hline
\textbf{Metric}&$\textbf{LAM}^1$ & $\textbf{LAM}^2$ & $\textbf{LAM}^3$&  \textbf{$\textbf{LAM}^4$} & \textbf{GPT-4o (Text-only)} &\textbf{GPT-4o Mini (Text-only)} \\ \hline
\textbf{Object Acc (\%)}  & 39.4 & 85.6 & 87.4 & \textbf{87.8}&73.2&74.6  \\ \hline
\textbf{Operation Acc (\%)} &59.9& 97.3 & 97.7 & \textbf{97.7}  &94.2&91.5 \\ \hline
\textbf{Status Acc (\%)}  &32.7 & 97.8 & 98.2 & \textbf{99.0} &52.1&67.4 \\ \hline
\textbf{Step Success Rate (SSR) (\%)} & 33.0& 83.6 & 85.9 & \textbf{86.2}&68.8& 73.4\\ \hline
\textbf{Task Success Rate (TSR) (\%)}  & 35.6& 76.8 & 79.3 & \textbf{81.2} & 67.2 &62.3\\ \hline
\end{tabular}
\label{tab:offline_eva}
\end{table*}

\subsubsection{Performance of \( \text{LAM}^1 \) on Planning}

Table~\ref{tab:task-plan-performance} presents the performance of \( \text{LAM}^1 \) in planning prediction across 15,334 tasks on Windows OS, utilizing the dataset detailed in Section~\ref{sec:task_plan}. \( \text{LAM}^1 \) achieves a TSR  of \textbf{82.2\%}, which is comparable to GPT-4o's TSR of \textbf{84.5\%}.
While GPT-4o demonstrates a slightly higher TSR, it exhibits a lower Step Precision of \textbf{28.2\%}, indicating inefficiencies in its planning by generating additional unnecessary steps. In contrast, \( \text{LAM}^1 \) achieves a higher Step Precision, reflecting its ability to produce more efficient and accurate plans. This superior precision is attributed to \( \text{LAM}^1 \)'s training regimen, which incorporates domain-specific knowledge through task-plan pretraining.

Additionally, the baseline Mistral-7B model, without any fine-tuning, performs inadequately with a TSR of \textbf{0.0\%}, Step Precision of \textbf{0.1\%}, and Step Recall of \textbf{0.5\%}. These stark results underscore the critical importance of task-plan pretraining in transforming a general-purpose language model into a competent task planner.

Overall, the evaluation highlights that while general-purpose models like GPT-4o can achieve high success rates, their lower step precision suggests a propensity for overcomplicating plans. In contrast, specialized models like \( \text{LAM}^1 \) not only maintain competitive success rates but also generate more streamlined and accurate action sequences. This validates the effectiveness of targeted training approaches in enhancing planning capabilities and demonstrates the necessity of task-plan pretraining for developing reliable and efficient task planners.


\subsection{Task-Action Results (Phases 2–4)}

\subsubsection{Evaluation Metrics}

To assess the performance of agents in completing tasks, we employ five primary metrics: \textbf{Object Accuracy (Object Acc.)}, \textbf{Operation Accuracy (Operation Acc.)}, \textbf{Status Accuracy (Status Acc.)}, \textbf{Step Success Rate (SSR)}, and \textbf{Task Success Rate (TSR)}. The definitions and calculation methods for these metrics are detailed below:

\begin{enumerate}[leftmargin=*]
    \item \textbf{Object Accuracy (Object Acc.):}  
    This metric measures the accuracy of selecting the correct control object for each task step. The predicted object is compared with the set of acceptable objects defined in the ground truth. It evaluates the agent's ability to correctly identify and interact with the appropriate UI elements.
    \item \textbf{Operation Accuracy (Operation Acc.):}  
    For operations such as \texttt{Click}, \texttt{Type}, or \texttt{Select Option}, this metric evaluates the correctness of the predicted action. It ensures that the agent performs the correct operation as specified in the ground truth.
    \item \textbf{Status Accuracy (Status Acc.):}  
    This metric assesses whether the agent correctly identifies the task's completion status based on its predictions. It evaluates the agent's understanding of the overall progression and whether the task is marked as finished appropriately.
    \item \textbf{Step Success Rate (SSR):}  
    A step is considered successful only if the selected object, predicted operation, and predicted status are all correct. This metric evaluates each step of the task independently by comparing the predicted outputs with the ground truth action history.
    \item \textbf{Task Success Rate (TSR):}  
    A task is considered successful only if all steps within the task are successful, making this a stringent evaluation metric. This metric provides a holistic measure of the agent's ability to complete complex, multi-step tasks accurately.
\end{enumerate}
These metrics collectively cover various aspects of agent performance, including precision in object selection, operation execution, task understanding, and overall task completion. By combining step-level and task-level evaluations, they provide a comprehensive assessment of the agent's effectiveness in real-world task execution.

\subsubsection{Performance on Decision Making}

Table~\ref{tab:offline_eva} summarizes the results on 435 tasks of the Word Application. The four-phase LAM training framework demonstrates incremental and cumulative improvements in task completion. Notably, \(\text{LAM}^4\) achieves a TSR of \(\mathbf{81.2\%}\), outperforming both GPT-4o (\(67.2\%\)) and GPT-4o-mini (\(62.3\%\)). This performance gap is substantial, considering that LAM’s training process relies on progressively collected data and incremental refinements tailored to each phase.

The step-by-step training strategy explains these gains. In Phase~1 (\(\text{LAM}^1\)), task-plan pretraining establishes a foundational understanding of task structures, resulting in a modest increase in TSR. In Phase 2 (\(\text{LAM}^2\)), imitation learning on GPT-4o-labeled success trajectories imparts efficient execution strategies, driving a significant jump in TSR from 35.6\% to 76.8\%. Phase 3 (\(\text{LAM}^3\)) introduces self-boosting exploration, where LAM autonomously tackles cases previously failed by GPT-4o. This yields an additional increase in TSR to 79.3\%. Finally, in Phase 4 (\(\text{LAM}^4\)), reward-guided fine-tuning refines decision-making based on sparse feedback, further elevating TSR to 81.2\%.

An important outcome is that the LAM framework enables the model to surpass GPT-4o, despite GPT-4o providing initial annotations. Through targeted data collection and progressive refinement, LAM not only assimilates the strengths of GPT-4o, but also learns from its failures to develop more robust and adaptable policies. The ReAct mechanism plays a crucial role here, allowing \(\text{LAM}^2\) and beyond to gather new success trajectories from challenging tasks, thereby enhancing its policy and overall performance.

In summary, the phased training approach and judicious data utilization enable LAM to excel where a state-of-the-art LLM (GPT-4o) falls short. This highlights the effectiveness of the LAM framework in crafting agents that are both data-efficient and capable of executing complex, multi-step tasks with high accuracy and reliability.

\section{Integration and Grounding\label{sec:grounding}}
Once the LAM is trained, we integrate it into the GUI agent UFO~\cite{ufo}, enabling the model's predicted actions to be grounded and executable within the Windows OS environment. The UFO agent accepts user requests in natural language and completes tasks by interacting with the UI controls of Windows applications. 

\subsection{LAM Agent In a Nutshell}
\begin{figure}[t]
    \centering
    \includegraphics[width=\columnwidth]{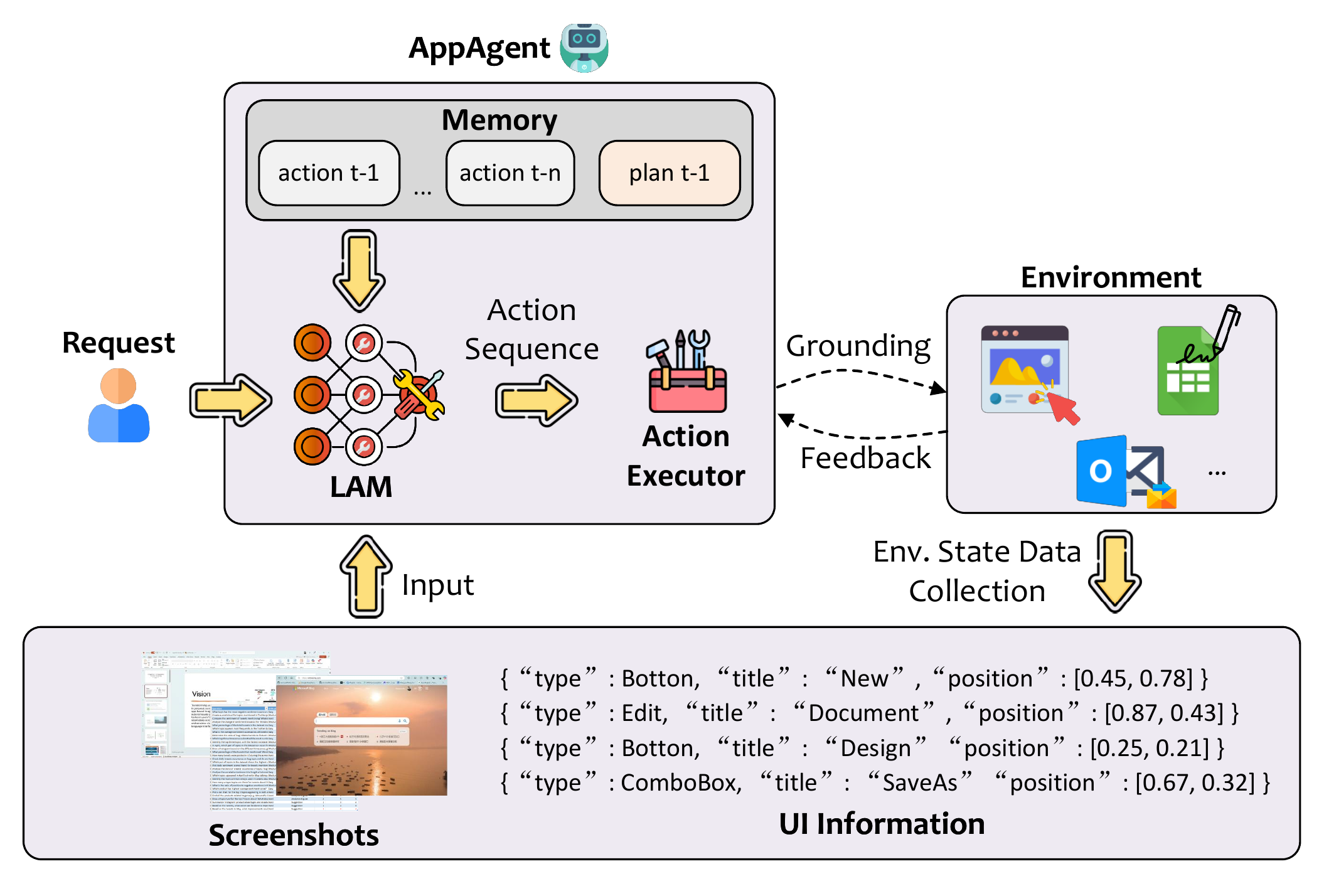}
    \vspace*{-2em}
    \caption{The overall architecture of the AppAgent employed in UFO.}
    \label{fig:appagent}
\end{figure}
In UFO, the LAM serves as the inference engine within the AppAgent, enabling efficient and accurate task completion. Figure~\ref{fig:appagent} illustrates the architecture of the AppAgent. UFO, equipped with LAMs, is designed for interactive engagement with Windows applications. For simplicity, we focus on automating tasks within Microsoft Word, a widely used productivity tool with a sophisticated GUI and diverse functionalities, making it an ideal testbed for training and evaluating LAM.

During each inference step, the agent collects critical contextual information from the application environment, which is then passed to the LAM for decision-making. The LAM performs planning, orchestrates actions, and infers the necessary steps to fulfill the user request. These inferred actions are grounded in the environment by mapping them to predefined tools and function calls used by the agent, such as mouse clicks, keyboard inputs, or API calls. This process iterates, with LAM continuously adjusting its plan based on real-time feedback from the environment, until the task is completed. Additionally, the agent maintains a memory that logs historical actions and plans, providing essential context for the LAM to make more informed and adaptive decisions as the task progresses. This integration ensures that UFO can efficiently manage and complete complex, real-world tasks in Windows environments.

\subsection{Environment}
The UFO agent leverages the LAM to interact with applications in the Windows environment. At each decision step, UFO employs the UI Automation (UIA) API \cite{dinh2018method} to inspect all actionable controls within the target Windows application, retrieving contextual information for each control\footnote{UIA is the native Windows OS APIs used to detect actionable controls and provide their metadata, such as names and locations. For other platforms, UIA can be replaced by vision-based detectors that analyze screenshots or by utilizing alternative accessibility APIs.}.
This information is passed to the LAM for control selection and action inference. The control data is structured as a list of dictionaries, where each control is assigned a numerical index (as a label), along with its title and control type, allowing the LAM to make informed decisions regarding control selection and the corresponding action. This input format mirrors the structure used during offline data collection for consistency in training and execution.


\subsection{LAM Inference}
Using the environmental observations of application control information, UFO constructs prompts in the same format as the offline training data, using planning and thought generation techniques~\cite{wei2022chain, ding2023everything} to enable LAM to make reliable inferences about the appropriate controls and operations to invoke. These inferences target the controls detected by the UIA, where each control is selected from a predefined list. The function calls inferred by LAM are limited to pre-defined operations, such as mouse and keyboard actions, as well as APIs specific to Word-related tasks. Once inferred, these operations are parsed and executed within the environment.

\subsection{Action Execution}
UFO employs a control interactor to ground the action strings generated by LAMs, translating them into tangible impacts within the target application. Each action typically consists of two key components:
\begin{enumerate}
    \item \textbf{Control Element:} This refers to the specific UI control within the application that will receive the action, such as a button, text box, or scroll bar.
    \item \textbf{Function Call:} This represents the operation to be performed on the control element, such as a mouse click, keyboard input, or invocation of native APIs.
\end{enumerate}
By combining the control element and its associated function call, UFO executes the inferred actions within the application.

\subsection{Memory}
UFO maintains additional information in its memory to assist LAMs in making more informed and accurate decisions. This memory includes:
\begin{enumerate}
    \item \textbf{Historical Actions:} A log of action trajectories and their execution results from the initial step onwards. This helps LAM understand the current system state and aids in exploring the next steps based on prior actions.    
    \item \textbf{Previous Plan:} The textual planning for future actions, generated by LAM in the previous step. This serves as a reference for guiding the current and future actions, ensuring consistency across steps.
\end{enumerate}
This memory is fed into LAM at each decision point, allowing for more effective decision-making. By maintaining a comprehensive record of past actions and plans, LAMs can better understand what has been accomplished, what remains to be done, and the outcomes of previous actions. This situational awareness enhances LAMs' ability to complete user requests more effectively and efficiently. 

\section{Online Evaluations\label{sec:online_eva}} 

With the integration of the Windows GUI agent UFO, we evaluate the performance of the LAM in real-world environments. The evaluation process and results are detailed in the following subsections.

\subsection{Testing Dataset}  
The online performance of LAM is evaluated on the same set of 435 test requests used during LAM training. The testing environments, specifically the Word document templates corresponding to each task, are also maintained as identical to the training setup to ensure consistency and comparability.

\begin{table*}[h]
\centering
\caption{Performance comparison of LAM and baseline models across metrics.}
\label{tab:online}
\begin{tabular}{l|c|c|c|c|c}
\hline
\multirow{2}{*}{\textbf{Metric}}  & \multicolumn{3}{c|}{\textbf{Text-only}}               & \multicolumn{2}{c}{\textbf{Text + Visual}} \\ \cline{2-6} 
                                  & \textbf{LAM} & \textbf{GPT-4o} & \textbf{GPT-4o Mini} & \textbf{GPT-4o}   & \textbf{GPT-4o Mini}   \\ \hline
\textbf{Task Success Rate (\%)}   & 71.0         & 63.0            & 57.8                 & 75.5              & 66.7                   \\ \hline
\textbf{Task Completion Time (s)} & 30.42        & 86.42           & 35.24                & 96.48             & 46.21                  \\ \hline
\textbf{Task Completion Steps}    & 5.62         & 6.73            & 5.99                 & 4.98              & 6.34                   \\ \hline
\textbf{Average Step Latency (s)} & 5.41         & 12.84           & 5.88                 & 19.36             & 7.29                   \\ \hline
\end{tabular}
\label{tab:model_performance}
\end{table*}

\subsection{Implementation}

Our LAM was deployed on a virtual machine (VM) configured as NC24s v3. The VM is equipped with 24 virtual cores (vCPUs), 448 GB of memory, and two NVIDIA Tesla V100 GPUs, each with 16 GB of memory, to support efficient inference. This computational setup was designed to meet the demanding requirements of LAM's inference processes effectively.

The UFO agent operates on six VMs running in parallel using Azure Dedicated Host\footnote{\url{https://azure.microsoft.com/en-us/products/virtual-machines/dedicated-host}} to accelerate the testing process. Each VM is equipped with a 15-core Intel(R) Xeon(R) Platinum 8370C CPU @ 2.80GHz, 64GB of RAM, and runs Windows 11 Enterprise version 23H2. Microsoft applications, such as Word and Excel, are installed on version 2410. GUI control is facilitated through the MSTSC tool\footnote{\url{https://learn.microsoft.com/en-us/windows-server/administration/windows-commands/mstsc}}. This setup ensures a consistent and controlled environment for evaluating the LAM's performance.

\subsection{Baselines}  
To benchmark the performance of LAM, we compared it against two baseline models: GPT-4o and GPT-4o Mini. These models are widely recognized for their robust natural language processing and reasoning capabilities, making them popular choices in the development of GUI agents. To ensure consistency in evaluation, the \texttt{top\_p} and \texttt{temperature} hyperparameters were set to 0 for both baseline models.

To further examine the impact of input modalities, we conducted an ablation study comparing performance with and without the inclusion of screenshots. Notably, LAM processes only textual inputs, excluding screenshots, while the baseline models were evaluated using both textual and visual modalities.

\subsection{Evaluation Metrics}  
We employ the following metrics to comprehensively evaluate the performance of LAM:

\begin{itemize}
    \item \textbf{Task Success Rate (TSR):} The percentage of tasks successfully completed out of the total tasks attempted. Task success is determined by an evaluation agent using GPT-4o, which assesses the full task completion trajectory, including plans, action sequences, and screenshots, to verify task completion.

    \item \textbf{Task Completion Time:} The total time taken to complete each task, measured from the initial request to the final action.

    \item \textbf{Task Completion Steps:} The total number of action steps performed by the agent to successfully complete each task.

    \item \textbf{Average Step Latency:} The average time taken per action step, reflecting the model's efficiency in generating and executing each action.

\end{itemize}

These metrics collectively evaluate both the accuracy and efficiency of task completion, providing a comprehensive assessment of the LAM's capabilities in real-world scenarios.



\subsection{Experimental Analysis}

The experimental results are presented in Table~\ref{tab:online}. LAM achieves a TSR of 71.0\%, demonstrating competitive performance compared to the GPT-4o models. While GPT-4o with visual inputs attains the highest TSR of 76.5\%, slightly outperforming LAM, its reliance on visual data introduces significant trade-offs in efficiency. Notably, when visual inputs are excluded, GPT-4o's TSR drops to 63.0\%, an 8.0 percentage point decrease compared to LAM. Similarly, GPT-4o Mini exhibits lower TSRs for both visual and non-visual settings (66.7\% and 57.8\%, respectively). These results underscore LAM's capability as a text-only model to maintain high task success rates, outperforming the text-only variants of the baseline models.

Efficiency is assessed through Task Completion Time and Average Step Latency, where LAM demonstrates clear superiority. LAM achieves the shortest Task Completion Time of \textbf{30.42 seconds}, substantially outperforming all baseline models. In comparison, GPT-4o without visual inputs records a completion time of 86.42 seconds, more than 2.84 times longer than LAM. GPT-4o with visual inputs fares even worse, with a completion time of 96.48 seconds. Although GPT-4o Mini models show slightly better efficiency than their larger counterparts, they remain less efficient than LAM, with completion times of 35.24 seconds (without visual inputs) and 46.21 seconds (with visual inputs).

LAM also excels in Average Step Latency, achieving the shortest time per action step at \textbf{5.41 seconds}. Without visual inputs, GPT-4o reduces its step latency to 12.84 seconds but still remains more than twice as slow as LAM. In comparison, GPT-4o with visual inputs exhibits the highest step latency at 19.36 seconds per step, more than triple LAM's latency.  GPT-4o Mini models show moderate improvements but still fall short, with step latencies of 7.29 seconds (with visual inputs) and 5.88 seconds (without visual inputs).

These findings highlight LAM's strengths as a text-only model, offering a compelling balance of competitive accuracy and superior efficiency. It achieves rapid task completion and low latency without sacrificing performance, making it an effective solution for real-world applications. Its specialized training enables precise action inference and execution, underscoring the potential of LAMs to enhance automation and productivity in agent-based systems.

\section{Limitation and Future Research}
While significant strides have been made in the development of LAMs, their current state is still in its infancy. Many technical challenges and limitations prevent LAMs from being fully productized and integrated into commercial use for real-world applications. Below, we outline key limitations and areas for future research to address these challenges.

\subsection{Safety Risk}
The ability of LAMs to perform real-world actions in physical or digital environments introduces significant safety risks. Unlike traditional LLMs, which primarily generate text, LAMs have the potential to manipulate external systems, control hardware, or make changes within software environments. While this capability is a key strength, it also presents a double-edged sword: errors in inference or execution can lead to unintended or harmful consequences \cite{liu2024towards, zhou2023making}.

For instance, a LAM controlling a robotic system could misinterpret a command and cause physical damage. Similarly, a LAM operating within a financial or healthcare application could execute erroneous actions with substantial real-world repercussions. Therefore, safety mechanisms such as formal verification, action validation, and fallback strategies must be integrated into LAM systems. Future research must focus on developing robust error detection, rollback mechanisms, and fail-safe systems that prevent actions from being executed until they have been thoroughly vetted for correctness and safety \cite{gehring1993neural, zhang2023safetybench, koo1987checkpointing}.

\subsection{Ethical and Regulatory Concerns}
The deployment of LAMs raises significant ethical and regulatory challenges \cite{biswas2023guardrails, yan2024practical, mesko2023imperative, minssen2023challenges, pineiro2023ethical}. As these models gain the ability to interact with real-world environments, questions about accountability, transparency, and fairness come to the forefront \cite{ferrara2024genai, liesenfeld2023opening, li2023survey}. For instance, who is held accountable if a LAM causes harm or damage due to a misinterpretation of a user's command? How do we ensure that these systems are making decisions in a fair and unbiased manner? These concerns are amplified by the fact that LAMs are often trained on large datasets that may contain biases, which can influence the model's decision-making processes \cite{navigli2023biases}.

Moreover, there are regulatory concerns regarding the deployment of LAMs in critical sectors such as healthcare, finance, and transportation, where strict guidelines govern the use of automated systems \cite{karabacak2023embracing, li2023large, cui2024receive}. Future research must address these concerns by developing transparent model architectures that allow for interpretability and explainability of actions taken by LAMs. Additionally, establishing clear regulatory frameworks and ethical guidelines will be crucial for ensuring that LAMs are deployed in a manner that prioritizes safety, fairness, and accountability.

\subsection{Scalability, Generalizability and Adaptability}
LAMs are often tailored to specific environments or scenarios, making their scalability, generalizability, and adaptability significant limitations. Most LAMs are designed to operate within a narrowly defined context, such as a specific operating system, application, or interface. These environments, however, are subject to frequent updates, changes in APIs, and the introduction of new applications or functionalities. A LAM trained on a specific version of an environment may fail when confronted with changes it has not encountered before, leading to poor performance or outright failures \cite{grosse2023studying, zhang2024out, kong2020calibrated}.

In addition, scaling LAMs to new environments or applications is challenging due to the high cost of collecting domain-specific data \cite{muennighoff2024scaling, minaee2024large}. Gathering sufficient training data for each new context is time-consuming and resource-intensive. Furthermore, the model's ability to generalize across different environments is often limited, as it may not be familiar with the nuances of new systems or tasks.

Future work should focus on improving the adaptability and generalizability of LAMs through techniques like transfer learning, multi-task learning, and few-shot learning. These approaches allow a model to generalize from one environment to another with minimal retraining. Moreover, developing automated data collection methods and self-supervised learning techniques could significantly reduce the effort required to scale LAMs to new domains.

\textbf{Summary.} While LAMs represent a promising advancement in the evolution of AI systems, they are still constrained by several technical, ethical, and practical limitations. Addressing these challenges will be essential for enabling the widespread adoption and commercialization of LAMs in real-world applications. By ensuring safety, addressing ethical concerns, and improving scalability and adaptability, future research can help unlock the full potential of LAMs.

\section{Related work}
The emergence of LAMs has led to significant impact across various agentic domains. In the following sections, we will review related research and practices at three levels: \emph{(i)} data of LAMs, \emph{(ii)} training LAMs, and \emph{(iii)} agents with LAMs.

\subsection{Data of LAMs}
The emergence of LLM-based agents has spurred the development of numerous datasets specifically tailored to LAM applications and their corresponding agent systems. These datasets can be divided into two main categories, namely \emph{(i)} \textbf{Datasets for Training LAMs:} These datasets provide the necessary input for training LAMs, including diverse user commands, environmental contexts, and action sequences. \emph{(ii)} \textbf{Evaluation Benchmarks:} These benchmarks are curated for testing and evaluating the capabilities of LAMs and agent systems. 

Mind2Web \cite{deng2024mind2web} is the first dataset developed for web agents that follow natural language instructions to complete complex tasks across diverse websites. It includes task descriptions, action sequences, and webpage snapshots, offering rich data for training and testing models in various web-based scenarios. Rawles \etal introduced a large dataset called Android in the Wild (AITW) \cite{rawles2024androidinthewild}, which is designed specifically for training models to control Android devices. SeeClick \cite{cheng2024seeclick} combines web, mobile, and general GUI tasks, creating a dataset of over 1 million samples for training LAMs. Similarly, GUICourse \cite{chen2024guicourse} and OmniACT \cite{kapoor2024omniact} provide datasets across web, smartphone, and desktop platforms, containing detailed user requests, environmental states, and action sequences. These datasets are invaluable resources for training LAMs in specific domains and evaluating their task execution abilities.

Several benchmarks have also been developed to evaluate the capabilities of LAMs and their associated agents in different environments. WebCanvas provides 542 tasks with dynamic environments, designed to assess the task completion ability of web agents. AndroidWorld \cite{rawles2024androidworld} offers a fully functional Android environment, featuring 116 programmatic tasks across 20 real-world Android apps with reward signals for performance evaluation. WindowsArena \cite{bonatti2024windowsagentarenaevaluating} focuses on benchmarking LAMs within the Windows GUI, while OSWorld \cite{xie2024osworld} extends this to a more diverse environment, encompassing Windows, macOS, and Ubuntu. These benchmarks provide standardized settings to measure and compare the effectiveness of LAMs and their agents in various real-world environments, enabling a unified evaluation framework for agentic models.

\subsection{Training LAMs}
Using both open and private domain-specific datasets, significant research efforts have been directed toward training LAMs for specialized purposes, enhancing the action inference abilities of traditional LLMs to enable automation and tangible real-world impact. For example, SeeClick \cite{cheng2024seeclick} and GUICourse \cite{chen2024guicourse}, in addition to releasing their own datasets, leverage these resources to train LAMs, grounding real-world data into models that effectively interact with their environments.

Hong \etal trained an 18-billion-parameter visual language LAM, named CogAgent \cite{hong2024cogagent}, which specializes in GUI understanding and navigation tasks across both PC and Android interfaces. By utilizing datasets like Mind2Web and AITW, CogAgent has been optimized for complex navigation and action execution tasks in diverse GUI environments. ScreenAI \cite{baechler2024screenai} introduced a textual representation for user interfaces (UIs) to teach models how to understand and interact with UIs. This approach also facilitates automatic generation of large-scale training data, which is then used to pretrain and fine-tune models for a wide spectrum of tasks, including UI and infographic understanding and navigation. Additionally, Zhang \etal released a series of large action models (xLAM) tailored for AI agent tasks \cite{zhang2024xlam}, including five models with both dense and mixture-of-expert architectures. By unifying datasets from diverse environments, xLAM ensures consistency in data format, simplifying model training and enhancing generalization across multiple benchmarks. These models have achieved outstanding performance in diverse scenarios, demonstrating the capability of LAMs to extend beyond traditional LLMs and perform complex real-world tasks.

These pioneering works have laid the foundation for advancing the action-oriented capabilities of LLMs, making LAMs a critical component in achieving robust automation and impactful real-world applications.

\subsection{Agents with LAMs}
With the development of LAMs, researchers have integrated these models into real-world agent systems, which provide the necessary components and workflows to ensure effective interaction between LAMs and their environments, enabling them to fulfill user requests efficiently. As a pioneer, Zhang \etal demonstrated that GPT-V can serve as a capable LAM for web navigation when coupled with appropriate agent techniques and tools, revealing the potential of LAMs in complex web interactions.
In the mobile domain, MobileAgent \cite{wang2024mobile} and AppAgent \cite{yang2023appagent} focus on automating tasks within Android applications by leveraging GUI agents. These systems demonstrate how LAMs can power task automation on mobile platforms, transforming how users interact with applications.

One of the most advanced systems, UFO \cite{ufo}, is a UI-focused agent designed for automating tasks on the Windows OS, further enhanced with APIs~\cite{lu2024turn}. UFO is composed of two key components: a HostAgent that decomposes user requests into subtasks and an AppAgent that executes these subtasks within individual applications. This architecture significantly enhances UFO's capability to handle cross-application tasks seamlessly, providing robust task automation across diverse software environments. 
In parallel, ScreenAgent \cite{niu2024screenagent}, Cradle \cite{tan2024towards}, OS-Copilot \cite{wu2024copilot}, and MMAC-Copilot \cite{song2024mmac} also focus on automating UI tasks in desktop environments. Notably, Cradle and OS-Copilot push the boundaries by enabling agents to learn from their experiences and self-evolve over time, further enhancing their effectiveness and autonomy.

By integrating LAMs into agents to handle complex tasks in these various scenarios, These pioneering efforts are opening new possibilities for the future of human-computer interaction, revolutionizing traditional methods of interacting with GUIs and paving the way for more intelligent, automated, and user-friendly systems.

\section{Conclusion}
``Actions speak louder than words.'' The transition from generating language responses to executing tangible actions marks the evolution of large language models into large action models, enabling them to make real-world impacts, a critical step towards achieving AGI. This technical report provides a comprehensive introduction to LAMs, covering their conceptual foundations, system architecture, and the step-by-step process of developing a LAM—from data collection to model training and deployment in real-world agent systems. We use the Windows OS environment and its GUI agent UFO, as a case study to demonstrate how to build a LAM from the ground up. Detailed implementation strategies and evaluation results are presented to offer practical insights into this process. 

However, despite progress, the development of high-quality LAMs is still in its early stages, with several limitations remaining. These include the extensive need for training data and computational resources, inference latency, and the risk of errors during real-world execution. While current LAMs have shown potential, there is substantial room for improvement. We anticipate that as these challenges are addressed, more sophisticated and reliable LAM applications will emerge, bringing us closer to fully autonomous systems capable of meaningful action in complex environments.

\bibliographystyle{acm}
\bibliography{reference}

\newpage

\appendix

\section{Template Word files}
\begin{figure}[H]
    \centering
    \includegraphics[width=\columnwidth]{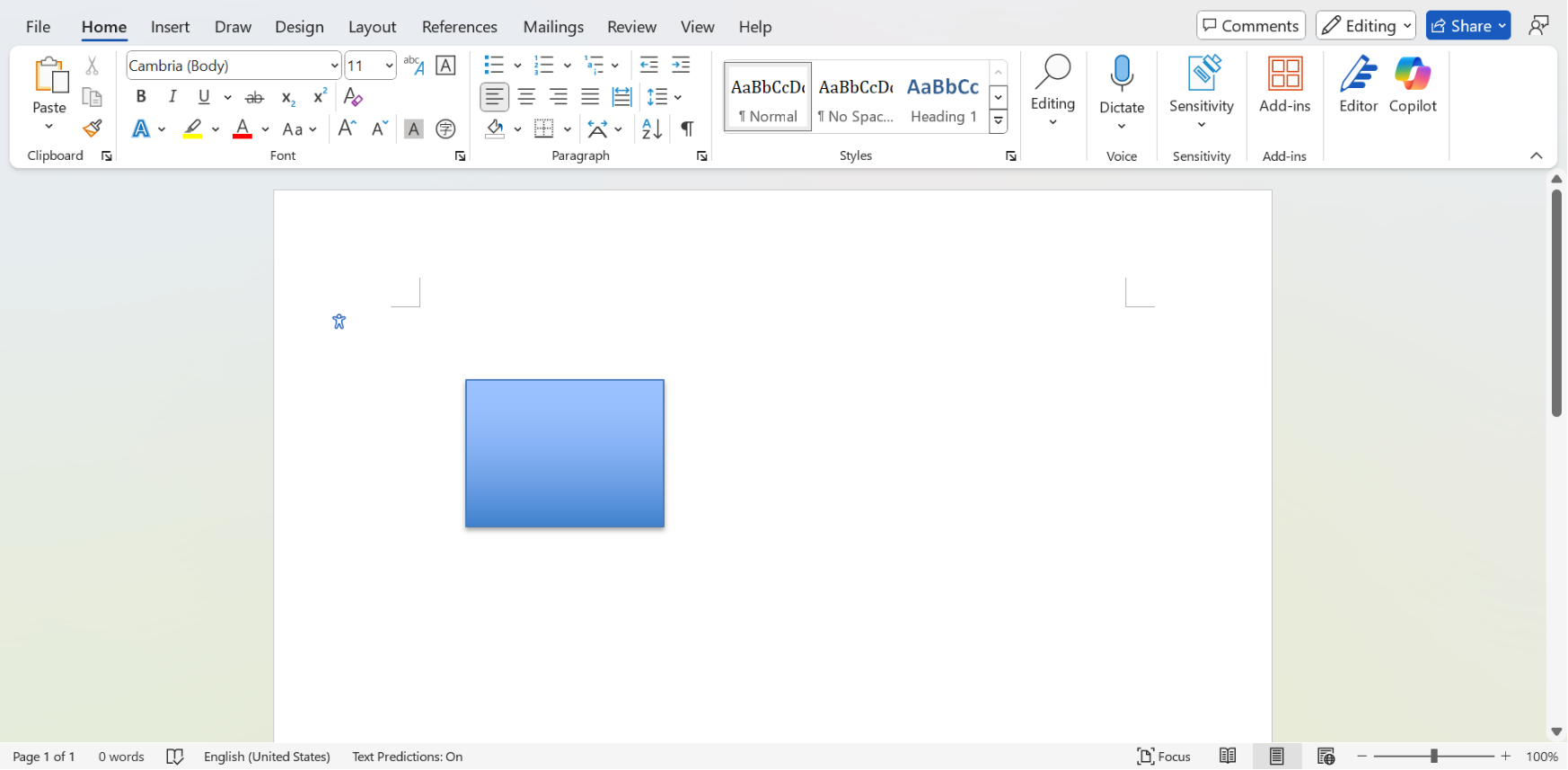}
    \caption{A word template file with the description ``A doc with a rectangle shape.''}
    \label{fig:tempalte1}
\end{figure}

\begin{figure}[H]
    \centering
    \includegraphics[width=\columnwidth]{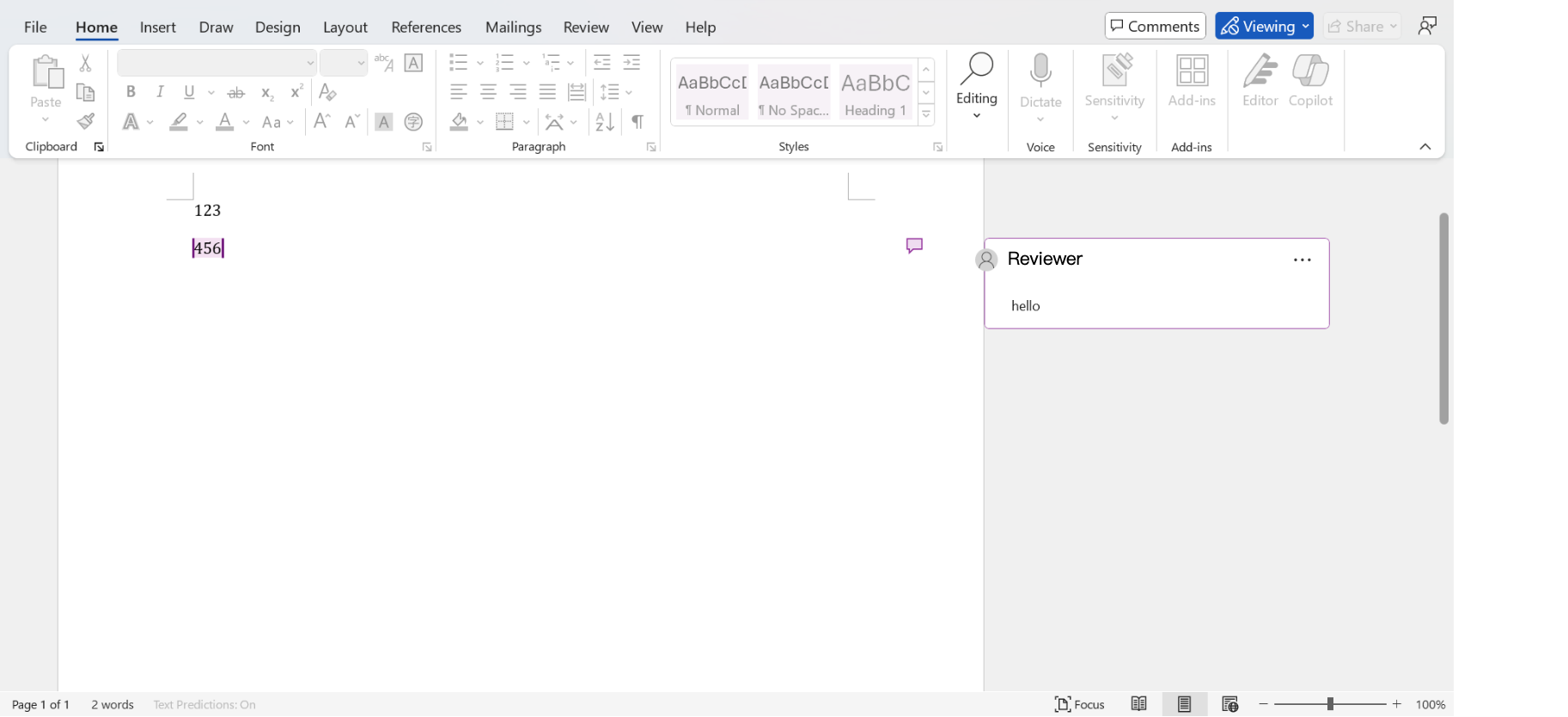}
    \caption{A word template file with the description ``A doc with comments and reviewer.''}
    \label{fig:tempalte2}
\end{figure}

\begin{figure}[H]
    \centering
    \includegraphics[width=\columnwidth]{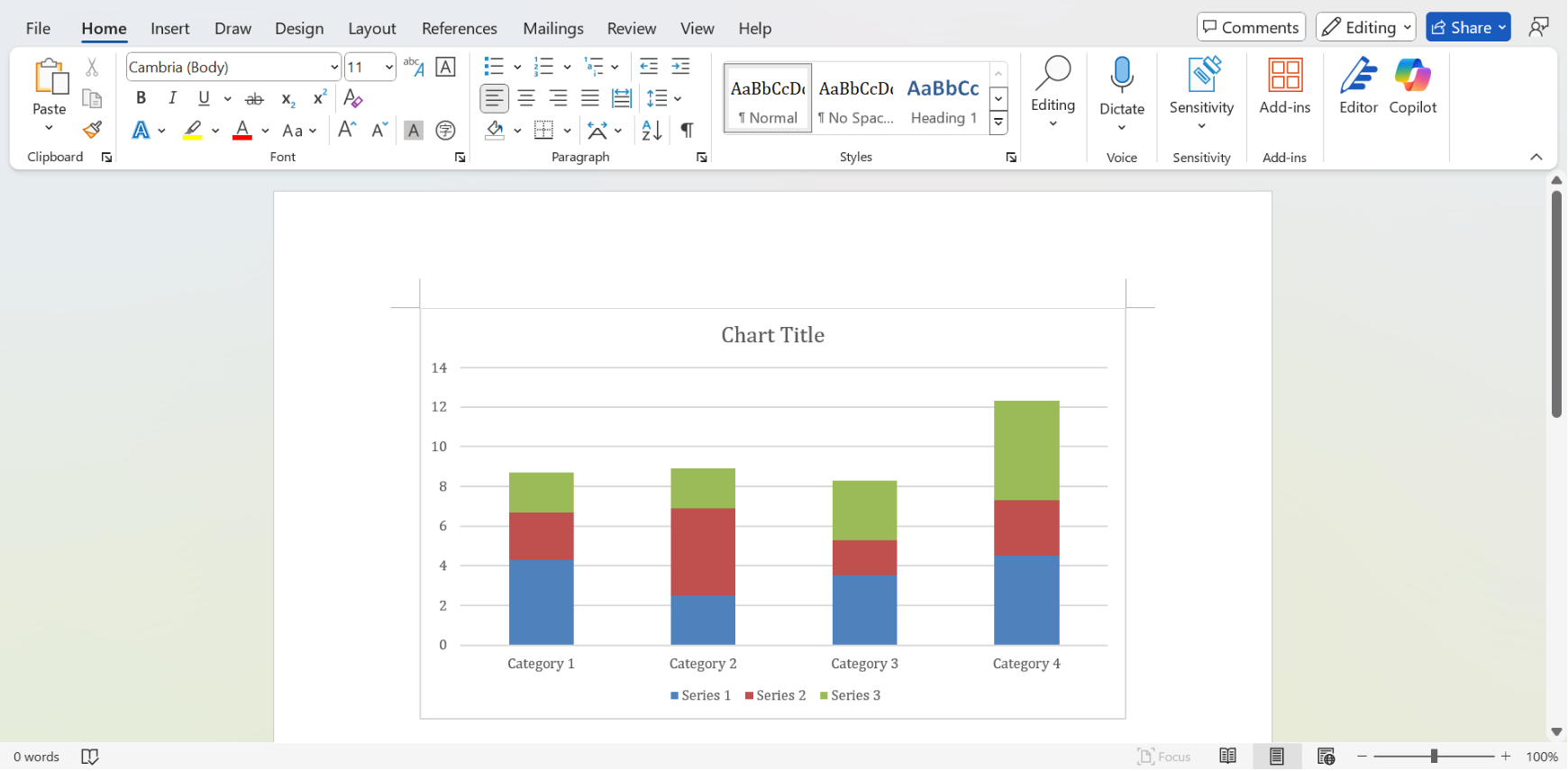}
    \caption{A word template file with the description ``A doc with a chart.''}
    \label{fig:tempalte3}
\end{figure}

Figure~\ref{fig:tempalte1}, ~\ref{fig:tempalte2}, and ~\ref{fig:tempalte3} show three template word file examples used in the instantiation phase when converting task-plan data to task-action data.
\label{append:templates-word}

\section{Prompts}
\subsection{Instantiation}
\label{append:instance}
The instantiation prompt used in the instantiation phase when converting task-plan data to task-action data.

\begin{lstlisting}[language=yaml]
system: |-
  You are a Agent Task Creator and planer.
  You will receive a <Given Task> that is abstract and your objective is to instantiate this task, and give the step-by-step actions to take.
  - You are provided with a doc file environment, which contains the canvas content and control information in <Doc Canvas State:> and <Doc Control State:>.
  - You should review the doc canvas content and control information to detail the <Given Task> to a <New Task>.The control information is in a dict tree of available control items format.
  - You are provided with <Available Actions>, you should review the acions carefully and choose the most suitable ones step-by-step <Action Plan>.
  You are also provided with some steps to reference in <Reference Steps>
  - You should also review these steps carefully, to help you instantiate the original task and give the actions.
  

  ## Control item
  - The control item is the element on the page that you can interact with, we limit the actionable control item to the following:
  - "Button" is the control item that you can click.
  - "Edit" is the control item that you can click and input text.
  - "TabItem" is the control item that you can click and switch to another page.
  - "ListItem" is the control item that you can click and select.
  - "MenuItem" is the control item that you can click and select.
  - "ScrollBar" is the control item that you can scroll.
  - "TreeItem" is the control item that you can click and select.
  - "Document" is the control item that you can click and select text.
  - "Hyperlink" is the control item that you can click and open a link.
  - "ComboBox" is the control item that you can click and input text. The Google search box is an example of ComboBox.

  ## Available Actions on the control item
  - All the available actions are listed below:
  {apis}

  ## The requirements for <New Task>
  1. The <New Task> must based on the given task.
  2. The <New Task> must be able to be completed step-by-step by a Windows Operating System or an Application on Windows platform.
  3. You should try your best not to make the <New Task> become verbose, <New Task> can only add up to 50 words into #Given Task#.
  4. The detailed target in <New Task> should be specific and clear based on the doc canvas content and control information.
  5. The <New Task> should be able to implemented by the available controls and actions.
  
  ## The requirements for <Action Plan>
  1. The <Action Plan> should be step-by-step actions to take in the doc file environment.
  2. Each action should be in the available actions from <Available Actions>.
  3. Each action should be generated with a "step" description which is the function description of the action.
  
  ## Response Format
  - You are required to response in a JSON format, consisting of several distinct parts with the following keys and corresponding content:
    {{
      "observation": <Outline the observation of the provided doc file environment based on the given Canvas State and Control State>,
      "thought": <Outline your thinking and logic of your New Task and the actions to take,consider the observation of environment and avaiable controls actions>,
      "new_task":<Give the detailed New Task based on Given Task and the observation of doc environment>,
      "actions_plan":<Give the detailed step-by-step actions plan based on the Available Actions and the observation of doc environment.,
      The format should be a list of action call format separated by "\n">
    }}
  
  ### Action Call Format
  - The action call format is the same as the available actions in the API list.You are required to provide the action call format in a JSON format:
    {{
    "step": <The step description the function of the action,which is also the subtask completed by the current action>
    "controlLabel": <Specify the precise annotated label of the control item to be selected, adhering strictly to the provided options in the field of "label" in the <Doc Control State:>. If you believe none of the control item is suitable for the task or the task is complete, kindly output a empty string ''.>
    "controlText": <Specify the precise control_text of the control item to be selected, adhering strictly to the provided options in the field of "control_text" in the <Doc Control State:>.The control text must match exactly with the selected control label. If the function to call do not need specify controlText or the task is complete,you can kindly output an empty string ''.
      If the function to call need to specify controlText and none of the control item is suitable for the task,you should input a possible control name.> 
      "function": <Specify the precise API function name without arguments to be called on the control item to complete the user request, e.g., click_input. Leave it a empty string "" if you believe none of the API function is suitable for the task or the task is complete.>
      "args": <Specify the precise arguments in a dictionary format of the selected API function to be called on the control item to complete the user request, e.g., {{"control_id":"1","button": "left", "double": false}}. Leave it a empty dictionary {{}} if you the API does not require arguments, or you believe none of the API function is suitable for the task, or the task is complete.>
    }}

    e.g.
      {{
          "step": "change the borders",
          "controlLabel": "",
          "controlText": "Borders",
          "function": "click_input",
          "args": {{
              "button": "left",
              "double": false
          }}
      }}

      {{
        "step": "change the borders",
          "controlLabel": "101",
          "controlText": "Borders",
          "function": "click_input",
          "args": {{
              "control_id": "101",
              "button": "left",
              "double": false
          }}
      }}

      {{
          "step": "select the target text",
          "controlLabel": "",
          "controlText": "",
          "function": "select_text",
          "args": {{
              "text": "Test For Fun"
          }}
      }}

  - The <actions_plan> field must be strictly in a format separated each action call by "\n". The list format should be like this: "action call 1\naction call 2\naction call 3"
  - If you think the original task do not need to be detailed, you can directly copy the original task to the "new_task".
  - You should review the apis function carefully and if the function to call need to specify target control,the "controlText" field
  cannot be set empty.
  - The "step" description should be consistent with the action and also the thought.

  ## Here are some examples for you to complete the user request:
  {examples}

  ## Tips
  - Read the above instruction carefully. Make sure the response and action strictly following these instruction and meet the user request.
  - Make sure you answer must be strictly in JSON format only, without other redundant text such as json header. Your output must be able to be able to be parsed by json.loads(). Otherwise, it will crash the system and destroy the computer.
  - Your task is very important to improve the agent performance. I will tip you 200$ if you do well. Thank you for your hard work!

user: |-
  <Given Task:> {given_task}
  <Reference Steps:> {reference_steps}
  <Doc Canvas State:> {doc_canvas_state}
  <Doc Control State:> {doc_control_state}
  <Your response:>
\end{lstlisting}

\subsection{Evaluation}
\label{append:evaluation}

The instantiation prompt used in the evaluation phase when converting task-plan data to task-action data.

\begin{lstlisting}[language=yaml]
system: |-
  You are an evaluator who can evaluate whether an agent has successfully completed a task in the <Original Request>. 
  The agent is an AI model that can interact with the desktop application and take actions. 
  The thought of agent plan is provided in the <Thought>.
  You will be provided with a task and the <Execution Trajectory> of the agent, including the agent actions that have been taken, and the change of environment. 
  You will also be provided with a final canvas state in <Final Env Status>.
  You will also be provided with a canvas difference in <Canvas Diff>.
  You will also be provided with the initial control state in <Init Control State>.
  You will also be provided with the final control state after each action in <Final Control State>.
  
  Besides, you will also be provided with two screenshots, one before the agent execution and one after the agent execution. 
  
  Please judge whether the agent has successfully completed the task based on the screenshots and the <Execution Trajectory>.You are required to judge whether the agent has finished the task or not by observing the screenshot differences and the intermediate steps of the agent.

  ## Execution trajectory information
  Here are the detailed information about a piece of agent execution trajectory item:
  - number: The number of action in the execution trajectory.
  - action: The action that the agent takes in the current step. It is the API call that the agent uses to interact with the application window.

  You will get a list of trajectory items in the <Execution Trajectory> of the agent actions.

  ### Control State
  
  - A control item is the element on the page that you can interact with, we limit the actionable control item to the following:
  - "Button" is the control item that you can click.
  - "Edit" is the control item that you can click and input text.
  - "TabItem" is the control item that you can click and switch to another page.
  - "ListItem" is the control item that you can click and select.
  - "MenuItem" is the control item that you can click and select.
  - "ScrollBar" is the control item that you can scroll.
  - "TreeItem" is the control item that you can click and select.
  - "Document" is the control item that you can click and select text.
  - "Hyperlink" is the control item that you can click and open a link.
  - "ComboBox" is the control item that you can click and input text. The Google search box is an example of ComboBox.
  - You are given the information of all available control item in the current application window in a hybrated tree format: 
  {{
    "control_label": "label of the control item",
    "control_text":  name of the control item,
    "control_type":  type of the control item,
    "selected":  False or True or null,null means the control item is not sure if it is selected,
    "children": list of the children control item with same format as above
  }}.

  ### Canvas State Format
  The canvas state is in the xml format which is transformed from the document object model (DOM) of the canvas area.
  The canvas diff is the difference of the canvas area before and after the action, which is in the format of the difference of the xml of the canvas area.
  Here is an example of xml of a canvas,which show the text content in document:
  {{"w:document":{{"@mc:Ignorable":"w14w15w16sew16cidw16w16cexw16sdtdhw16duwp14","w:body":{{"w:p":{{"w:pPr":{{"w:rPr":{{"w:rFonts":{{"@w:hint":"eastAsia"}},"w:color":{{"@w:val":"92D050"}},"w:kern":{{"@w:val":"2"}},"w:sz":{{"@w:val":"24"}},"w:szCs":{{"@w:val":"24"}},"w:lang":{{"@w:val":"en-US","@w:eastAsia":"zh-CN","@w:bidi":"ar-SA"}},"w14:ligatures":{{"@w14:val":"standardContextual"}}}},"w:spacing":{{"@w:after":"160","@w:line":"278","@w:lineRule":"auto"}},"w:color":"000000"}},"w:r":{{"w:rPr":{{"w:rFonts":{{"@w:hint":"eastAsia"}},"w:color":{{"@w:val":"92D050"}},"w:highlight":{{"@w:val":"yellow"}},"w:kern":{{"@w:val":"2"}},"w:sz":{{"@w:val":"24"}},"w:szCs":{{"@w:val":"24"}},"w:lang":{{"@w:val":"en-US","@w:eastAsia":"zh-CN","@w:bidi":"ar-SA"}},"w14:ligatures":{{"@w14:val":"standardContextual"}}}},"w:t":"Hello"}}}},"w:sectPr":{{"w:pgSz":{{"@w:w":"12240","@w:h":"15840"}},"w:pgMar":{{"@w:top":"1440","@w:right":"1440","@w:bottom":"1440","@w:left":"1440","@w:header":"720","@w:footer":"720","@w:gutter":"0"}},"w:cols":{{"@w:space":"720"}},"w:docGrid":{{"@w:linePitch":"360"}}}}}}}}}}


  ### Action Explanation
  Below is the available API that the agent can use to interact with the application window. You can refer to the API usage to understand the agent actions.
  {apis}

  ## Evaluation Items

  You have 2 main items to evaluate:

  1. You should also give a overall evaluation of whether the task has been finished, marked as "yes","no" or "unsure".
  2. You should also give a overall evaluation of the quality of task,marked as "ambiguous","over-detailed" or "good".

  Criteria for evaluation of the task completion:
  1. The <Final Control State:> and <Final Env Status:> should be consistent with the task requirements.If the 
  controls or canvas content expected to be changed are not changed, the task is not completed.
  2. The <Execution Trajectory> should be consistent with the task requirements. If the agent actions are not consistent with the task requirements, the task is not completed.
  3. If any action in the <Execution Trajectory> is empty, the task is not completed.



  Criteria for evaluation of the task quality:
  1. The description of the <Original Request:> should be clear and unambiguous, without the meaning of "selection".
  2. The description of the <Original Request:> should not be too detailed like step-by-step actions.
  
  ## Response Format

  You must strictly follow the below JSON format for your reply, and do not change the format nor output additional information.
  {{
      "task_quality": The quality of the <Original Request:>, which is "ambiguous/over-detailed/good",
      "task_complete": The evaluation of the task completion, which is "yes/no/unsure",
      "complete_judgement": your judgment of whether the task has been finished, and the detailed reasons for your judgment based on the provided information,
      "quality_judgement": your judgment of the quality of the task, and the detailed reasons for your judgment based on the provided information
  }}

  Please take a deep breath and think step by step. Observe the information carefully and analyze the agent execution trajectory, do not miss any minor details. 
  Rethink your response before submitting it.
  Your judgment is very important to improve the agent performance. I will tip you 200$ if you provide a detailed, correct and high-quality evaluation. Thank you for your hard work!
  
user: |-
  <Original Request:> {request}
  <Thought:> {thought}
  <Execution Trajectory:> {trajectory}
  <Canvas Diff:> {canvas_diff}
  <Init Control State:> {init_control_state}
  <Final Control State:> {final_control_state}
  <Final Env Status:> {final_status}
  <Your response:>
\end{lstlisting}

\section{Templates of training format}
\label{sec:template}
The following presents a template of the training data format. The parts enclosed in ``{}'' are fields that need to be filled. The ``apis'' field corresponds to the function information in the respective app, while ``control\_item'' contains the control information of the app under the current screenshot. The ``user\_request'' field captures the user's current request, ``step\_history'' records the agent's previous trajectory history, and ``previous\_plan'' outlines the agent's planning for the task in the previous state.

\begin{lstlisting}[language=yaml]
system: |-
  - You are a virtual assistant that can help users to complete their current requests by interacting with the UI of Window OS.
  - You are provided a list of control items of the current application window for reference
  - You are provided your previous plan of action for reference to decide the next step,the previous plan is the list of plan for the future actions made before the current action.
  - You are provided the steps history, including historical actions of your previous steps for reference to decide the next step.
  - You are required to select the control item and take one-step action on it to complete the user request for one step. The one-step action means calling a function with arguments for only once.
  - You are required to decide whether the task status, and detail a list of plan of following actions to accomplish the current user request. Do not include any additional actions beyond the completion of the current task.

  ## Control item
  - The control item is the element on the page that you can interact with, we limit the actionable control item to the following:
  - "Button" is the control item that you can click.
  - "Edit" is the control item that you can click and input text.
  - "TabItem" is the control item that you can click and switch to another page.
  - "ListItem" is the control item that you can click and select.
  - "MenuItem" is the control item that you can click and select.
  - "ScrollBar" is the control item that you can scroll.
  - "TreeItem" is the control item that you can click and select.
  - "Document" is the control item that you can click and select text.
  - "Hyperlink" is the control item that you can click and open a link.
  - "ComboBox" is the control item that you can click and input text.

  ## Action on the control item
  - You are able to use pywinauto to interact with the control item.
  {apis}


  ## Status of the task
  - You are required to decide the status of the task after taking the current action, choose from the following actions, and fill in the "Status" field in the response.
    - "CONTINUE": means the task is not finished and need further action.
    - "FINISH": means the current task is finished for the AppAgent and no further actions are required.

  ## Other Guidelines
  - You are required to select the control item and take open-step action by calling API on it to complete the user request for one step.
  - You are required to response in a JSON format, consisting of 7 distinct parts with the following keys and corresponding content:
    {{
    "thought": <Outline your thinking and logic of current one-step action required to fulfill the given request. You are restricted to provide you thought for only one step action.>
    "control_label": <Specify the precise annotated label of the control item to be selected, adhering strictly to the provided options in the field of "label" in the control information. If you believe none of the control item is suitable for the task or the task is complete, kindly output a empty string ''.>
    "control_name": <Specify the precise control_text of the control item to be selected, adhering strictly to the provided options in the field of "control_text" in the control information. If you believe none of the control item is suitable for the task or the task is complete, kindly output a empty string ''. The control text must match exactly with the selected control label.>
    "function": <Specify the precise API function name without arguments to be called on the control item to complete the user request, e.g., click_input. Leave it a empty string "" if you believe none of the API function is suitable for the task or the task is complete.>
    "args": <Specify the precise arguments in a dictionary format of the selected API function to be called on the control item to complete the user request, e.g., {{"button": "left", "double": false}}. Leave it a empty dictionary {{}} if you the API does not require arguments, or you believe none of the API function is suitable for the task, or the task is complete.>
    "status": <Specify the status of the task given the action.>
    "plan": <Specify the following list of plan of action to complete the user request. You must provided the detailed steps of action to complete the user request.If you believe the task is finished and no further actions are required after the current action, leave it an empty list.>
    }}

user: |-
  <Available Control Item:> {control_item}
  <User Request:> {user_request}
  <Previous Actions:> {step_history}
  <Previous Plans:> {previous_plan}

assistant: |-
  {output}
\end{lstlisting}

\section{Evaluation Prompt for Task-Plan}\label{app:task-plan-eval}

The evaluation prompt for results from \(\text{LAM}^1\) after task-plan pretraining.

\begin{lstlisting}[language=yaml]
You are a helpful and precise assistant for checking the quality of the answer. We would like to invite you to evaluate the performance of two AI assistants in answering a user's question in <Question>. These two answers are in <Answer1> and <Answer2>, respectively. Your evaluation will contain five sub-evaluation tasks:
1. Can <Answer1> solve the user's question?
    - Your answer should be "Yes" or "No".
2. Can <Answer2> solve the user's question?
    - Your answer should be "Yes" or "No".
3. Both two answers contain a list of steps marked by numbers. Your task is to extract action items from the provided steps in both answers. The action item is defined like a combination of action and element. Compare the action items to identify similarities. Output the similar action items. Count the count of similar action items.
    - Your answer should contain the extracted two action item sets (in the format as a list of string).
    - Your answer should contain the set of similar action items (in the format as a list of string). Similar action items are those sharing similar intent or achieving similar goals. Each similar action pair in the list should be in the format of "similar action item from action item set1 / similar action item from action item set2"
    - Your answer should contain the count of similar action items.
4. Which assistant provides a more helpful response?
    - Your answer should be "1" or "2", where "1" represents <Answer1> and "2" represents <Answer2>.
    - Your answer should contain the reason(s) for your choice. You should not focus on the length of the answer or the details of the answer, but you should focus on whether the steps could solve the user's question and the quality of the steps.

Your output should be in the following format in json:
{{
    "Subtask1": "Yes" or "No",
    "Subtask2": "Yes" or "No",
    "Subtask3": {{
        "Action items in Answer1": ["action item 1", "action item 2", ...],
        "Action items in Answer2": ["action item 1", "action item 2", ...],
        "Similar action items": ["similar action item 1", "similar action item 2", ...],
        "Count of similar action items": 2
    }},
    "Subtask4": {{
        "More helpful assistant": "1" or "2",
        "Reason": "reason for your choice"
    }}
}}

Here is the user's question <Question>: {question}
The first answer <Answer1> is: {answer1}
The second answer <Answer2> is: {answer2}

\end{lstlisting}

\section{LAM Training Objectives \label{app:obj}}
The problem is formally structured into two key objectives: \textit{(i)} task-plan pretraining and \textit{(ii)} decision-making training.

{\it Task-plan pretraining} aims to enable the LAM to map a given task description to a structured sequence of plans necessary for accomplishing the task. The primary objective of this component is to generate accurate and coherent plans. The training dataset consists of task-plan pairs, defined as:
\[
\mathcal{D}_{\text{plan}} = \{(t_i, P_i)\}_{i=1}^N
\]
where $t_i$: The task description, $P_i$: A sequence of plans to complete the task.

In {\it decision-making training}, the dataset consists of task-action trajectories, defined as::
\[
\tau = \{(s_1, a_1), (s_2, a_2), \dots, (s_T, a_T)\}
\]
where:
\begin{itemize}[leftmargin=*]
    \item $s_t$ (state at time step $t$), comprising:
    \begin{itemize}[leftmargin=*]
        \item \textbf{Task description}: A high-level summary of the task.
        \item \textbf{Step ID}: The current step in the task sequence.
        \item \textbf{Observations}: Information including control elements and the current canvas state.
        \item \textbf{Thoughts}: Model-generated reasoning for the current step.
        \item \textbf{Previous actions and plans}: The sequence of actions and plans from prior steps.
    \end{itemize}
    \item $a_t$ (action taken at time step $t$), consisting of:
    \begin{itemize}[leftmargin=*]
        \item \textbf{Thought}: Model's reasoning for the action.
        \item \textbf{Control label}: A label for the control element.
        \item \textbf{Control name}: The name of the control to interact with.
        \item \textbf{Function name}: The specific function invoked by the action.
        \item \textbf{Arguments}: Parameters passed to the function.
        \item \textbf{Status}: Indicates action's progress, either ongoing (\textit{Continue}) or completed (\textit{Finish}).
    \end{itemize}
\end{itemize}
The objective of decision-making training is to train the LAM to predict the appropriate action $a_t$ for a given state $s_t$ at each time step. This enables the model to map input states to corresponding actions across the sequence of steps required to accomplish the task.

\end{document}